# MEGA-CL: A Molecular Foundation Model for Generalizable ADMET Prediction through Graph External Attention and Contrastive Learning

*Tinghui Jin[1,#], Kedu Jin[1,#], Ying Li[1], Guanghui Ren[1], Jingzhi Xue[1], Shiyu Zhou[1], Xiaoli Dai[2], Libin Wei[3], Xijing Chen[1,*], Di Zhao[1,*], Jinfeng Liu[1,*]*

*[1]School of Science, School of Basic Medicine and Clinical Pharmacy, China Pharmaceutical University, Nanjing, China*

*[2]Sir Run Run Hospital, Nanjing Medical University, Nanjing, China*

*[3]Jiangsu Key Laboratory of Carcinogenesis and Intervention, China Pharmaceutical University, Nanjing, China*

*To whom correspondence should be addressed: jinfengliu@cpu.edu.cn (J.L.), zh_d99@cpu.edu.cn (D.Z.), chenxj-lab@hotmail.com (X.C,)

[#]These authors contributed equally to this work

## Abstract

Predicting the absorption, distribution, metabolism, excretion and toxicity (ADMET) properties of small molecules remains a major challenge in drug discovery, largely due to the limited size, heterogeneity and experimental noise of available datasets. Here, we present MEGA-CL, a foundation graph neural network framework for universal molecular ADMET prediction, pre-trained on approximately 100 million unlabeled molecules and subsequently fine-tuned on task-specific datasets. MEGA-CL integrates self-supervised contrastive learning with a multi-head external attention mechanism and an enhanced message-passing architecture, enabling simultaneous modeling of local chemical substructures and global inter-graph relationships while mitigating over-smoothing effects commonly observed in deep graph networks. Across 13 benchmark datasets and 21 downstream ADMET tasks, MEGA-CL consistently outperforms state-of-the-art baseline models. In particular, the framework demonstrates robust performance on challenging regression tasks, including clearance (CL) and steady-state volume of distribution (VDss), while maintaining strong generalization ability in independent external validation. Clinically relevant predictive accuracy was achieved, with more than 75% of predictions falling within a 3-fold error range. In an external evaluation on 18 novel compounds derived from recently approved FDA drugs, over 50% of human liver microsome clearance (HLMC) predictions were within a 2-fold error range. To further assess its practical applicability, MEGA-CL was prospectively evaluated on three preclinical drug candidates using in vitro hepatic microsomal metabolism assays and CYP450 inhibition assays guided by model predictions. The predicted HLMC values for all candidates were within 2.5-fold of the experimentally measured values, and 73.3% of CYP450 inhibition endpoints (11/15) were correctly classified. These results demonstrate the potential of MEGA-CL as a generalizable framework for accelerating in silico ADMET evaluation and early-stage drug candidate optimization.

## Introduction

The efficient discovery and development of safe and effective therapeutic agents remain central objectives of modern pharmaceutical research[1–3]. Pharmacokinetics (PK), which characterizes the ADMET processes of drugs in vivo, plays a critical role in bridging early-stage molecular design and clinical evaluation[4]. Accurate characterization of PK properties is essential for understanding drug exposure, optimizing dosing regimens, and assessing potential safety liabilities. However, suboptimal PK and toxicity profiles continue to represent major causes of drug attrition, accounting for approximately 10-15% of clinical trial failures and resulting in substantial economic and temporal costs[4,5]. Conventional PK evaluation relies heavily on labor-intensive, costly, and low-throughput in vitro and in vivo assays[6,7]. Moreover, empirical interspecies scaling methods used to extrapolate preclinical data to humans often introduce considerable predictive uncertainty, thereby limiting translational reliability from animal models to clinical trials[8,9].

To overcome the limitations of experimental PK and ADMET evaluation, computational approaches have become increasingly important in early-stage drug discovery[10,11]. In particular, advances in machine learning have enabled quantitative structure-activity relationship (QSAR) methods to predict molecular properties directly from chemical structures[12,13]. Among these approaches, graph neural networks (GNNs) have emerged as powerful frameworks for molecular representation learning due to their ability to naturally encode graph-structured chemical information[14–16]. Nevertheless, the performance and generalizability of supervised learning models, including both conventional machine learning approaches (e.g., SVM, XGBoost) and GNN-based architectures, remain constrained by the limited scale and variable quality of experimentally annotated ADMET datasets[17–19]. As a result, models trained on relatively small labeled datasets often exhibit poor robustness and limited transferability across the vast and diverse chemical space encountered in practical drug discovery applications.[12,17,20]

Recent advances in self-supervised learning have provided promising strategies for alleviating the dependence on large-scale annotated molecular datasets[21–23]. In particular, contrastive learning frameworks such as MolCLR leverage graph-based augmentations to pre-train molecular encoders on millions of unlabeled compounds, thereby enabling the extraction of transferable chemical representations[21,24]. Despite these advances, most existing approaches rely on conventional message-passing GNNs, which often suffer from limited expressive capacity, over-smoothing effects, and reduced ability to capture long-range atomic interactions as network depth increases[25,26]. In addition, current molecular representation learning methods predominantly focus on local intra-graph structural information while largely neglecting higher-order relationships across molecules within the broader chemical space[21,27]. Intuitively, molecules sharing similar scaffolds or substructural motifs exhibit correlated physicochemical properties[28]. The lack of explicit modeling of such inter-molecular dependencies may therefore limit the generalizability and predictive performance of existing frameworks for complex ADMET and PK-related tasks[24,27].

To address these limitations, we developed MEGA-CL, a large-scale pre-trained GNN framework for molecular ADMET prediction. MEGA-CL is trained using a self-supervised contrastive learning strategy that enables the model to learn transferable molecular representations from large collections of unlabeled compounds[29]. Architecturally, the framework combines an enhanced message-passing backbone (GCN+) with a graph external attention (GEA) mechanism to jointly model local intra-molecular features and higher-order relationships across molecules. Specifically, the GCN+ architecture improves the expressiveness of molecular representations while alleviating over-smoothing effects commonly observed in deep GNNs[30,31]. In parallel, the multi-head GEA module captures global inter-molecular correlations within the broader chemical space, thereby facilitating the learning of shared physicochemical and pharmacokinetic patterns across structurally related compounds[21,32–34]. Across multiple benchmark datasets and independent external evaluations, MEGA-CL consistently outperforms established machine learning methods and state-of-the-art deep learning baselines. These results demonstrate the potential of MEGA-CL as a generalizable computational framework for accelerating preclinical ADMET evaluation and drug candidate optimization.

## Results and Discussions

### Framework of MEGA-CL

The MEGA-CL framework integrates self-supervised contrastive learning with a multi-head GEA mechanism for molecular representation learning. As illustrated in Fig. 1A, each molecular SMILES string $s_n$ is first converted into a molecular graph $G_n$. Two correlated graph views are subsequently generated through stochastic graph augmentation, yielding $\tilde{G}_{2n-1}$ and $\tilde{G}_{2n-1}$. These augmented graphs are processed by a shared encoder composed of an enhanced graph convolution (GCN+) backbone and a multi-head GEA module to produce graph-level representations $h$, which are further projected into latent vectors $z$ through a multilayer perceptron (MLP) projection head.

During pre-training, the normalized temperature-scaled cross-entropy (NT-Xent) loss is used to maximize representation agreement between positive molecular pairs while contrasting them against other molecules within the same batch. Three graph augmentation strategies are used, including atom masking, bond deletion, and subgraph removal, as illustrated in Fig. 1B. The GCN+ backbone incorporates residual connections and layer normalization to enable deeper message passing while alleviating over-smoothing effects. In parallel, the multi-head GEA module captures higher-order correlations across molecules to refine graph-level representations. As shown in Fig. 1C, MEGA-CL is pre-trained on approximately 100 million unlabelled chemical molecules using self-supervised contrastive learning, with the pre-training loss curve shown in Supplementary Fig. S1. The pre-trained encoder is subsequently transferred to downstream ADMET prediction tasks and fine-tuned using task-specific prediction heads.

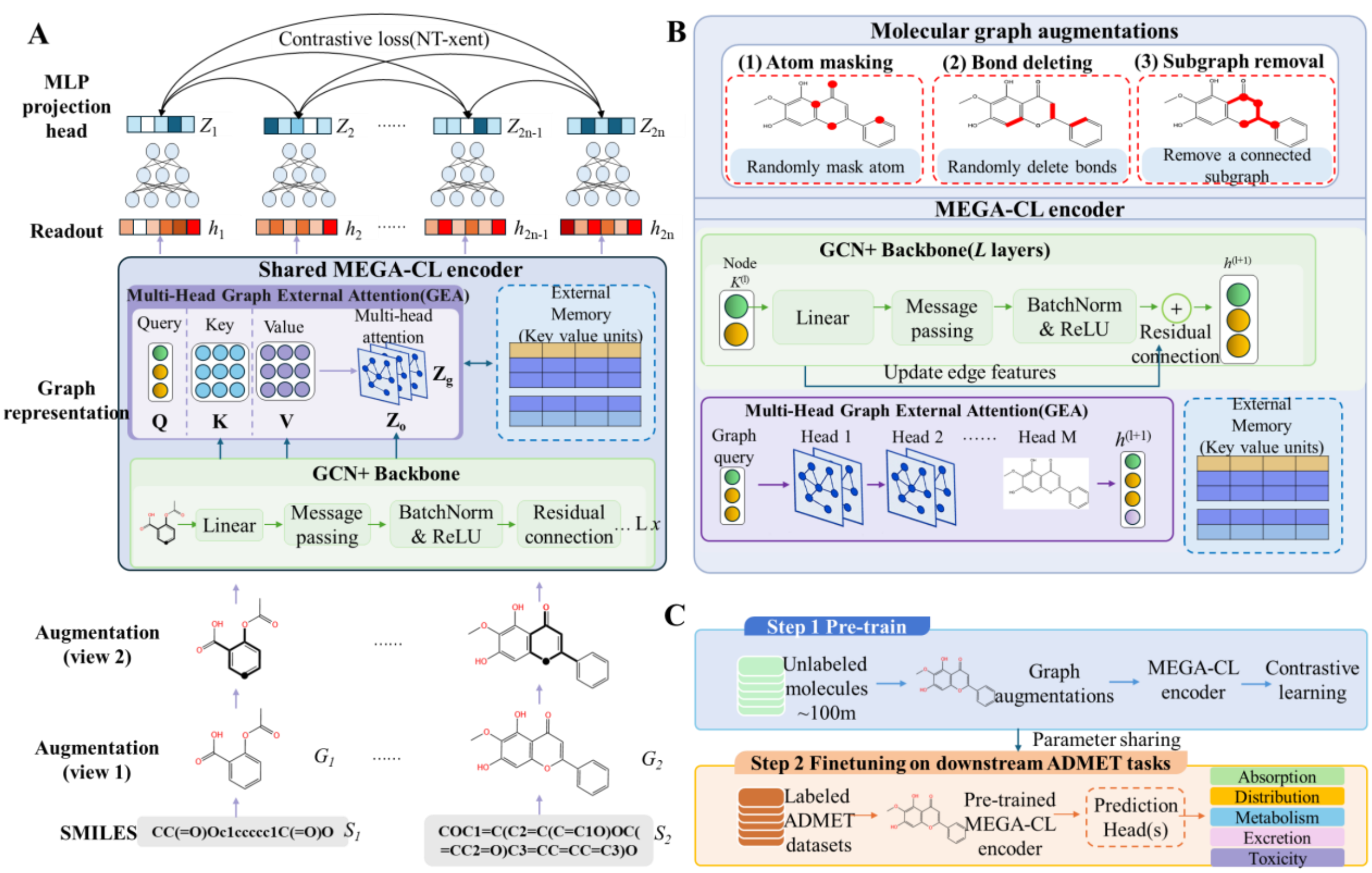

**Fig. 1 Overview of the MEGA-CL framework. A.** Contrastive pre-training pipeline of MEGA-CL. Molecular SMILES strings are converted into molecular graphs, followed by stochastic graph augmentations to generate correlated graph views. The augmented graphs are encoded by a shared MEGA-CL encoder consisting of a GCN+ backbone and a multi-head GEA module to produce graph-level representations, which are further projected into latent vectors through an MLP projection head. The NT-Xent loss is used to maximize agreement between positive pairs while contrasting against other molecules within the batch. **B.** Molecular graph augmentation strategies and architecture of the MEGA-CL encoder. Three augmentation strategies are employed during pre-training, including atom masking, bond deletion, and subgraph removal. The encoder integrates an enhanced graph convolution backbone with residual connections and a multi-head GEA module to capture both local intra-molecular features and higher-order inter-molecular

correlations. **C.** Finetuning for downstream ADMET prediction. MEGA-CL is first pre-trained on approximately 100 million unlabeled molecules using self-supervised contrastive learning and subsequently fine-tuned on labeled downstream ADMET datasets with task-specific prediction heads.

## Benchmark performance on molecular property prediction tasks

After pre-training, MEGA-CL was fine-tuned on 7 classification and 6 regression tasks from MoleculeNet to evaluate benchmark performance. These tasks cover diverse drug physicochemical and PK properties, as well as quantum mechanical properties (QM7/8/9), which have been widely used as benchmarks in various studies[15,21,35]. The benchmark results on MoleculeNet classification and regression tasks are summarized in Tables 1 and 2, respectively. Overall, MEGA-CL showed strong and consistent performance across molecular property prediction tasks, achieving the best or second-best performance on 4 of 7 classification benchmarks and 5 of 6 regression benchmarks. Notably, the performance gains are particularly pronounced in regression tasks associated with continuous physicochemical and quantum mechanical property prediction.

**Table 1. ROC-AUC on classification benchmark tasks**

| Dataset | BBBP | Tox21 | ClinTox | HIV | BACE | SIDER | MUV |
|---|---|---|---|---|---|---|---|
| **RF** | 71.4 ± 0.0 | 76.9 ± 1.5 | 71.3 ± 5.6 | 78.1 ± 0.6 | 86.7 ± 0.8 | **68.4 ± 0.9** | 63.2 ± 2.3 |
| **SVM** | 72.9 ± 0.0 | **81.8 ± 1.0** | 66.9 ± 9.2 | **79.2 ± 0.0** | 86.2 ± 0.0 | 68.2 ± 1.3 | 67.3 ± 1.3 |
| **GCN** | 71.8 ± 0.9 | 70.9 ± 2.6 | 62.5 ± 2.8 | 74.0 ± 3.0 | 71.6 ± 2.0 | 53.6 ± 3.2 | 71.6 ± 4.0 |
| **GIN** | 65.8 ± 4.5 | 74.0 ± 0.8 | 58.0 ± 4.4 | 75.3 ± 1.9 | 70.1 ± 5.4 | 57.3 ± 1.6 | 71.8 ± 2.5 |
| **SchNet**[15] | 84.8 ± 2.2 | 77.2 ± 2.3 | 71.5 ± 3.7 | 70.2 ± 3.4 | 76.6 ± 1.1 | 53.9 ± 3.7 | 71.3 ± 3.0 |
| **MGCN**[36] | 85.0 ± 6.4 | 70.7 ± 1.6 | 63.4 ± 4.2 | 73.8 ± 1.6 | 73.4 ± 3.0 | 55.2 ± 1.8 | 70.2 ± 3.4 |
| **D-MPNN**[37] | 71.2 ± 3.8 | 68.9 ± 1.3 | 90.5 ± 5.3 | 75.0 ± 2.1 | 85.3 ± 5.3 | 63.2 ± 2.3 | **76.2 ± 2.8** |
| **Hu et al.**[38] | 70.8 ± 1.5 | 78.7 ± 0.4 | 78.9 ± 2.4 | 80.2 ± 0.9 | 85.9 ± 0.8 | 65.2 ± 0.9 | 81.4 ± 2.0 |
| **N-Gram**[39] | **91.2 ± 3.0** | 76.9 ± 2.7 | 85.5 ± 3.7 | **83.0 ± 1.3** | **87.6 ± 3.5** | 63.2 ± 0.5 | 81.6 ± 1.9 |
| **MolCLR$_{GCN}$**[21] | 73.8 ± 0.2 | 74.7 ± 0.8 | 86.7 ± 1.0 | 77.8 ± 0.5 | 78.8 ± 0.5 | 66.9 ± 1.2 | 84.0 ± 1.8 |
| **MolCLR$_{GIN}$** | 73.6 ± 0.5 | **79.8 ± 0.7** | **93.2 ± 1.7** | 80.6 ± 1.1 | **89.0 ± 0.3** | **68.0 ± 1.1** | **88.6 ± 2.2** |
| **MEGA-CL** | **90.4 ± 0.7** | **81.2 ± 0.9** | **93.37 ± 0.7** | 80.81 ± 0.71 | 86.6 ± 1.7 | **69.1 ± 0.6** | 87 ± 0.5 |

Mean ± SD over three repeat runs. Two best results are in bold.

**Table 2 RMSE and MAE on regression benchmark tasks**

| Dataset | FreeSolv | ESOL | Lipo | QM7[a] | QM8[a] | QM9[a] |
|---|---|---|---|---|---|---|
| **SVM** | 3.14 ± 0.00 | 1.50 ± 0.00 | 0.82 ± 0.00 | 156.9 ± 0.0 | 0.0543 ± 0.0010 | 24.613 ± 0.144 |
| **GCN** | 2.87 ± 0.14 | 1.43 ± 0.05 | 0.85 ± 0.08 | 122.9 ± 2.2 | 0.0366 ± 0.0011 | 5.796 ± 1.969 |
| **GIN** | 2.76 ± 0.18 | 1.45 ± 0.02 | 0.85 ± 0.07 | 124.8 ± 0.7 | 0.0371 ± 0.0009 | 4.741 ± 0.912 |
| **SchNet**[15] | 3.22 ± 0.76 | 1.05 ± 0.06 | 0.91 ± 0.10 | **74.2 ± 6.0** | 0.0204 ± 0.0021 | **0.081 ± 0.001** |
| **MGCN**[36] | 3.35 ± 0.01 | 1.27 ± 0.15 | 1.11 ± 0.04 | 77.6 ± 4.7 | 0.0223 ± 0.0021 | **0.050 ± 0.002** |
| **D-MPNN**[37] | **2.18 ± 0.91** | **0.98 ± 0.26** | **0.65 ± 0.05** | 105.8 ± 13.2 | **0.0143 ± 0.0022** | 3.241 ± 0.119 |
| **Hu et al.**[38] | 2.83 ± 0.12 | 1.22 ± 0.02 | 0.74 ± 0.00 | 110.2 ± 6.4 | 0.0191 ± 0.0003 | 4.349 ± 0.061 |
| **N-Gram**[39] | 2.51 ± 0.19 | 1.10 ± 0.03 | 0.88 ± 0.12 | 125.6 ± 1.5 | 0.0320 ± 0.0032 | 7.636 ± 0.027 |
| **MolCLR$_{GCN}$**[21] | 2.39 ± 0.14 | 1.16 ± 0.00 | 0.78 ± 0.01 | 83.1 ± 4.0 | 0.0181 ± 0.0002 | 3.552 ± 0.041 |
| **MolCLR$_{GIN}$** | 2.20 ± 0.20 | 1.11 ± 0.01 | 0.65 ± 0.08 | 87.2 ± 2.0 | 0.0174 ± 0.0013 | 2.357 ± 0.118 |
| **MEGA-CL** | **1.40 ± 0.28** | **0.98 ± 0.07** | **0.63 ± 0.03** | **68.3 ± 8.6** | **0.0139 ± 0.0002** | 4.407 ± 0.137 |

Mean ± SD over three repeat runs. Two best results are in bold. [a] Measured by MAE.

MEGA-CL maintained competitive performance even on relatively small datasets, suggesting improved robustness under limited-data conditions. For example, on the FreeSolv[40]dataset containing only 642 molecules, MEGA-CL reduced the prediction error by approximately 36% compared with the second-best D-MPNN model[37]. Similar trends were observed across multiple benchmark datasets when compared with existing pre-trained and self-supervised models[15,21,36,37,39,41], suggesting that large-scale self-supervised pre-training on unlabeled molecules improves representation transferability and downstream generalization.

Compared with existing conventional contrastive learning frameworks like MolCLR series models, MEGA-CL achieved improved performance on 5 out of 7 classification benchmarks and reduced prediction errors on 5 of 6 regression tasks. These improvements are likely attributable to the complementary integration of local structural encoding and global relational representation learning. The enhanced GCN+ backbone facilitates deeper and more stable message passing while mitigating over-smoothing effects, thereby preserving fine-grained molecular structural information. Meanwhile, the multi-head GEA mechanism captures higher-order correlations among molecules within the broader chemical space, enabling the model to learn transferable physicochemical patterns beyond individual graph topology. Together, these components provide a more expressive molecular representation framework for diverse ADMET and physicochemical property prediction tasks. Having established competitive performance on standard molecular property benchmarks, we next evaluated MEGA-CL on a curated ADMET benchmark designed to reflect practical preclinical screening scenarios.

## Construction of a comprehensive ADMET benchmark

To systematically evaluate the practical utility of MEGA-CL in drug discovery, we curated a comprehensive ADMET benchmark comprising 21 downstream tasks that span the major stages of drug absorption, distribution, metabolism, excretion, and toxicity. Dataset construction prioritized physiological relevance, data quality, and sufficient sample size (>1,000 compounds per task whenever possible), with human-derived experimental measurements preferentially retained to maximize translational value.

The benchmark includes 14 classification tasks and 7 regression tasks. Metabolism-related endpoints comprise CYP450 inhibitor and substrate prediction across five major isoforms together with human liver microsomal clearance (HLM). Drug absorption and transport are represented by P-gp substrate recognition, PAMPA permeability, and Caco-2 permeability, whereas toxicity assessment includes Ames mutagenicity and drug-induced liver injury (DIList). In addition, five clinically relevant pharmacokinetic parameters (Fu,

VDss, CL, MRT, and HL) were incorporated as regression tasks to evaluate the model's capability for continuous-value prediction.

Following data collection, all molecular structures were standardized using unified SMILES representations and subjected to quality control, deduplication and label harmonization. The resulting benchmark provides a diverse and challenging evaluation platform for assessing molecular representation learning methods across a broad spectrum of ADMET endpoints. Detailed dataset composition, task types, and units are summarized in Supplementary Table S1. The distribution of regression and classification datasets is shown in Supplementary Fig. S2 and S3, respectively.

## Performance across diverse ADMET endpoints

To further evaluate the practical utility of MEGA-CL in drug discovery, we assessed its performance on a curated ADMET benchmark comprising 21 downstream tasks (14 classification and 7 regression tasks) spanning absorption, distribution, metabolism, excretion, and toxicity endpoints. The benchmark includes both classification and regression tasks and covers a diverse range of experimentally measured pharmacological and pharmacokinetic properties. The results for all classification and regression tasks are reported in Supplementary Table S2 and S3. As shown in Fig. 2, MEGA-CL consistently achieved competitive or superior performance across most ADMET prediction tasks compared with conventional machine learning methods and state-of-the-art graph-based models including Random forest (RF)[42], Support vector machine (SVM)[43], eXtreme Gradient Boosting (XGBoost)[44], Logistics classification[45], Naïve Bayes and the Directed Message Passing Neural Networks (D-MPNNs) from Chemprop[46]. Particularly notable improvements were observed for metabolism-related endpoints. For CYP450 inhibitor prediction in Fig. 2A, MEGA-CL achieved the highest ROC–AUC values on four of the five major isoforms (CYP1A2, CYP2C19, CYP2D6, and CYP3A4) and remained competitive on CYP2C9. Similar trends were observed for CYP450 substrate prediction in Fig. 2B, where MEGA-CL consistently ranked among the top-performing models across all evaluated isoforms. These results suggest that the learned molecular representations effectively capture structural features associated with enzyme recognition and metabolic liability.

MEGA-CL also demonstrated strong performance on absorption- and toxicity-related tasks as shown in Fig. 2C. For transporter and permeability prediction, including P-gp substrate recognition and PAMPA permeability, the model achieved performance comparable to or exceeding that of existing approaches. Likewise, competitive accuracy was observed on toxicity-related benchmarks such as AMES mutagenicity and drug-induced liver injury (DIList), indicating that the learned representations generalize across distinct biological mechanisms and experimental assay types.

More pronounced advantages were observed for continuous pharmacokinetic property prediction as shown in Fig. 2D. Across six regression tasks, including fraction unbound in plasma (Fu), human liver microsomal clearance (HLM), steady-state volume of distribution (VDss), mean residence time (MRT), systemic clearance (CL), and half-life (HL), MEGA-CL consistently achieved lower prediction errors than most competing methods. The improvements were particularly evident for clinically relevant PK parameters such as CL and VDss, which are known to be challenging due to the complex interplay between molecular structure, metabolism, tissue distribution, and protein binding. For Caco-2 permeability, all present methods achieved very low prediction errors (with RMSEs $< 0.5$), though the error of MEGA-CL is slightly larger than the competing methods. These results indicate that MEGA-CL is capable of capturing subtle physicochemical determinants underlying drug disposition and pharmacokinetic behavior.

The strong performance across diverse ADMET endpoints is likely attributable to the complementary integration of local structural encoding and global relational representation learning within the MEGA-CL framework. The enhanced GCN+ backbone enables effective extraction of fine-grained molecular features while mitigating over-smoothing during message passing. Meanwhile, the multi-head graph external attention module captures higher-order relationships among molecules within the broader chemical space, facilitating the transfer of shared pharmacological and physicochemical knowledge across structurally related compounds. Together, these components provide a more expressive and transferable molecular representation that supports robust prediction across a wide range of ADMET tasks.

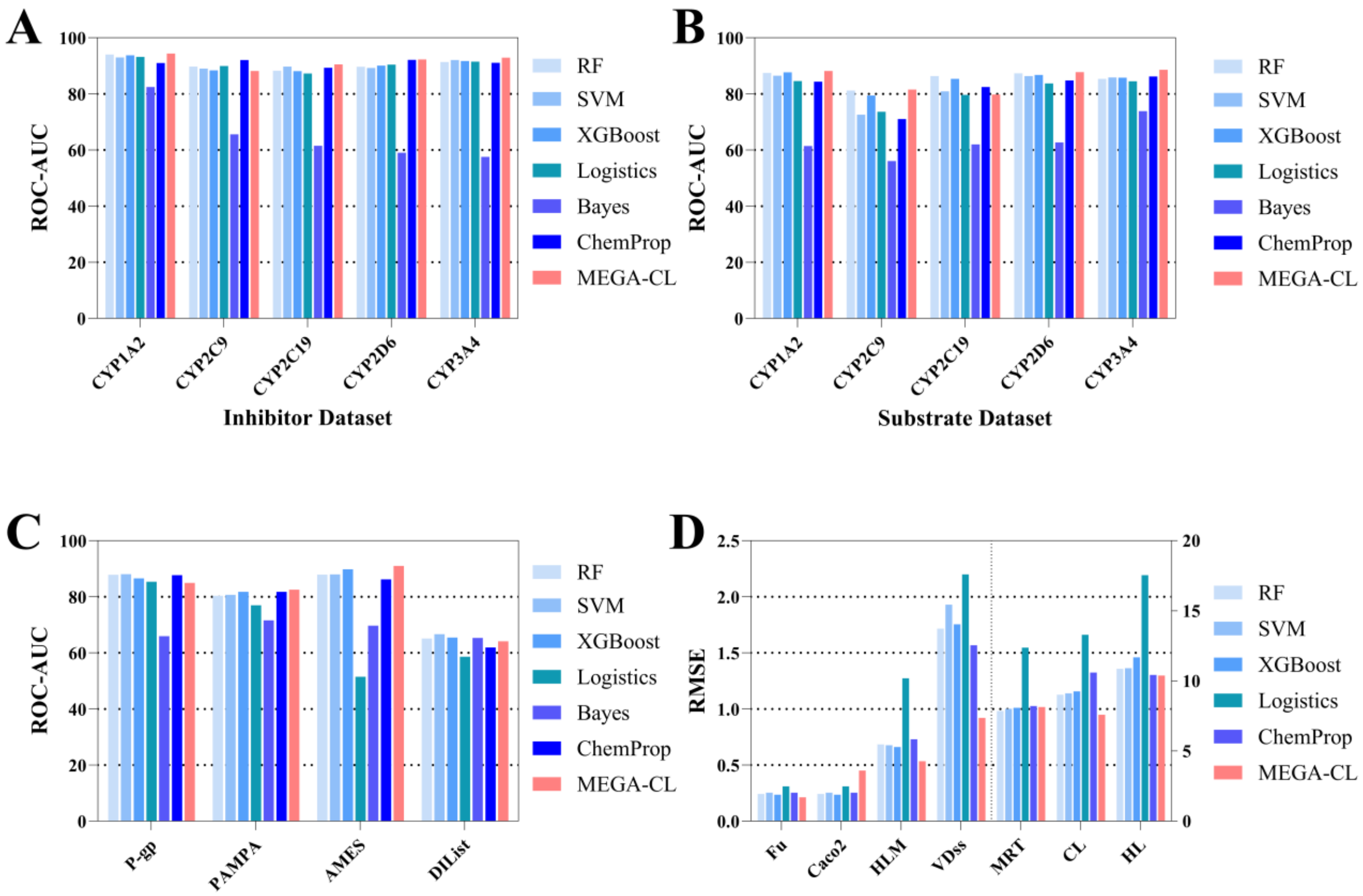

**Fig. 2 Performance of MEGA-CL across metabolism, transport, toxicity, and pharmacokinetic ADMET endpoints. A**, ROC–AUC for CYP450 inhibitor prediction across five major isoforms. **B**, ROC–AUC for CYP450 substrate prediction. **C**, ROC–AUC for transport and toxicity endpoints (P-gp, PAMPA, AMES, and DIList). **D**, RMSE for pharmacokinetic regression tasks (Fu, Caco-2, HLM, VDss, MRT, CL, and HL). Bars represent the mean over three independent runs of 100 epochs.

## Accurate prediction of continuous pharmacokinetic properties

MEGA-CL demonstrates strong performance in continuous pharmacokinetic and ADMET regression tasks, which require fine-grained modeling of structure–property relationships and are typically more sensitive to data noise and biological variability than classification tasks. Across all evaluated endpoints as shown in Fig. 3A, the model achieves consistently well-controlled fold-error distributions, with more than 75% of predictions falling within a 5-fold error range for each dataset. Importantly, performance is not uniform across tasks, but instead reflects clear stratification according to underlying pharmacokinetic complexity.

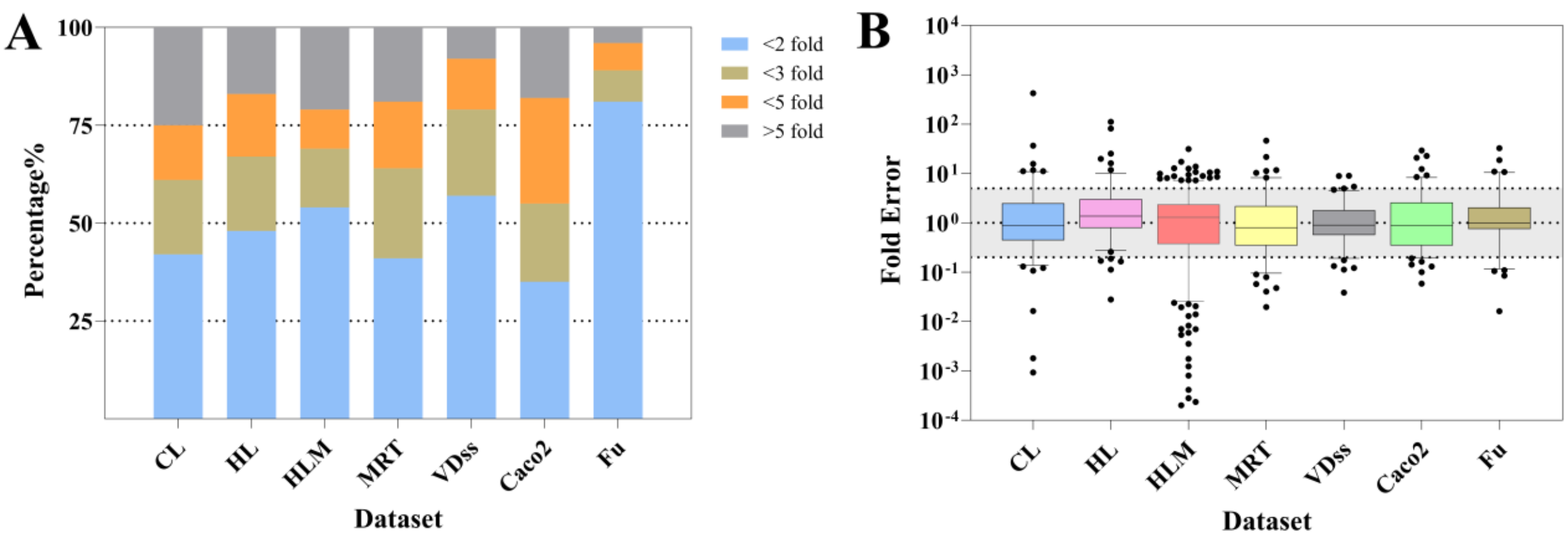

**Fig. 3 Fold-error analysis of continuous pharmacokinetic and ADMET prediction tasks. A.** Distribution of fold-error proportions across regression datasets, including clearance (CL), fraction unbound in plasma (Fu), half-life (HL), human liver microsomal clearance (HLM), mean residence time (MRT), steady-state volume of distribution (VDss), and Caco-2 permeability. Predictions are categorized into four relative error ranges (<2-, <3-, <5-, and >5-fold), illustrating overall prediction reliability across pharmacokinetic endpoints. **B.** Distribution of relative fold deviations for each dataset. Box-and-whisker plots summarize prediction errors on a logarithmic scale, highlighting task-dependent variability in continuous ADMET prediction. The gray shaded region indicates a 5-fold error range, and extreme deviations are shown as outliers.

For structure-driven properties such as fraction unbound in plasma (Fu), MEGA-CL achieves the most accurate predictions, with more than 75% of predictions within a 2-fold error range. This suggests that protein-binding–related endpoints are predominantly governed by local molecular features, which can be effectively captured by learned structural representations. In contrast, system-level pharmacokinetic parameters, including clearance (CL), half-life (HL), and mean residence time (MRT), exhibit broader error distributions and increased sensitivity to extreme deviations, with a higher proportion of predictions falling outside the 5-fold range. These endpoints are known to emerge from the integration of multiple physiological processes, including metabolic turnover, tissue distribution, and elimination pathways, making them inherently more difficult to model from molecular structure alone. Intermediate endpoints such as human liver microsomal clearance (HLM) and steady-state volume of distribution (VDss) show balanced behavior between these two extremes, with over 50% of predictions were within a 2-fold error range, reflecting partial dependence on both molecular properties and biological system-level factors.

To further characterize the distribution of relative deviations, we showed the full fold-error distributions across datasets in Fig. 3B. The results also reveal distinct task-dependent patterns. Caco-2 permeability exhibited relatively concentrated error distributions, with approximately 82% of predictions within a 5-fold range but only moderate performance within 2-fold deviation (35%). HLM and CL, while achieving mod-erate proportions of predictions within 5-fold error, exhibited heavier tails and more frequent outliers, reflecting their sensitivity to biological variability and experimental noise. In contrast, Fu demonstrated both compact distributions and low extreme-error frequency, further confirming its strong structure dependence.

Notably, despite their complexity, MEGA-CL maintains stable predictive performance across these tasks, indicating that the learned representations capture both local chemical determinants and higher-order pharmacokinetic dependencies. Overall, these results suggest that MEGA-CL is not only capable of predicting continuous ADMET properties with competitive accuracy, but also exhibits a meaningful degree of task-aware generalization across different levels of pharmacokinetic complexity. This supports its potential utility for quantitative pharmacokinetic modeling in early-stage drug discovery.

## External prospective experimental validation of MEGA-CL

To further assess the practical applicability of MEGA-CL in real-world drug discovery scenarios, we conducted a comprehensive external prospective experimental validation using preclinical compounds. Three structurally diverse preclinical compounds were selected, including two natural products (Oroxylin A and Icariin) and one synthetic compound (Compound B). Their phase I metabolic stability and CYP450 inhibition profiles were experimentally characterized using human liver microsomal incubation assays and standard CYP450 inhibition assays. Species-specific metabolic kinetic parameters, compound-level predictions, and external validation summaries are provided in Supplementary Table S4, S5 and S6, respectively.

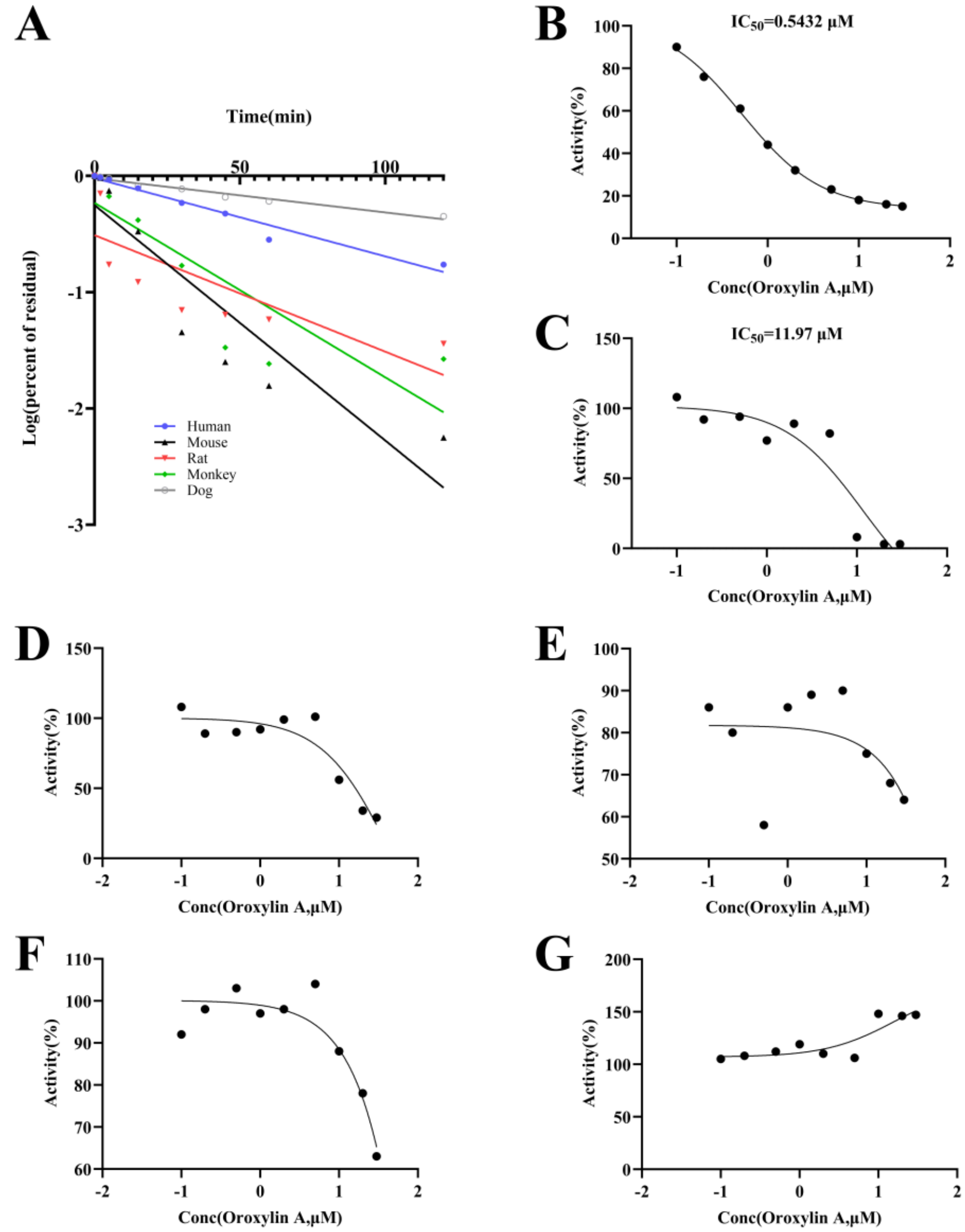

**Fig. 4 In vitro metabolic characterization and CYP450 inhibition profiles of oroxylin A. A.** Metabolic stability of oroxylin A in liver microsomes from multiple species, including human, rat, mouse, dog, and monkey, illustrating interspecies differences in metabolic clearance profiles. **B-G.** In vitro CYP450 inhibition profiles of oroxylin A against major drug-metabolizing enzymes. Panels B-G show its effects on probe substrate metabolism mediated by CYP1A2 (phenacetin), CYP2C9 (diclofenac), CYP2C19 (S-Mephenytoin), CYP2D6 (dextromethorphan), and CYP3A4 (testosterone and midazolam, representing CYP3A4 C2 and M activity, respectively).

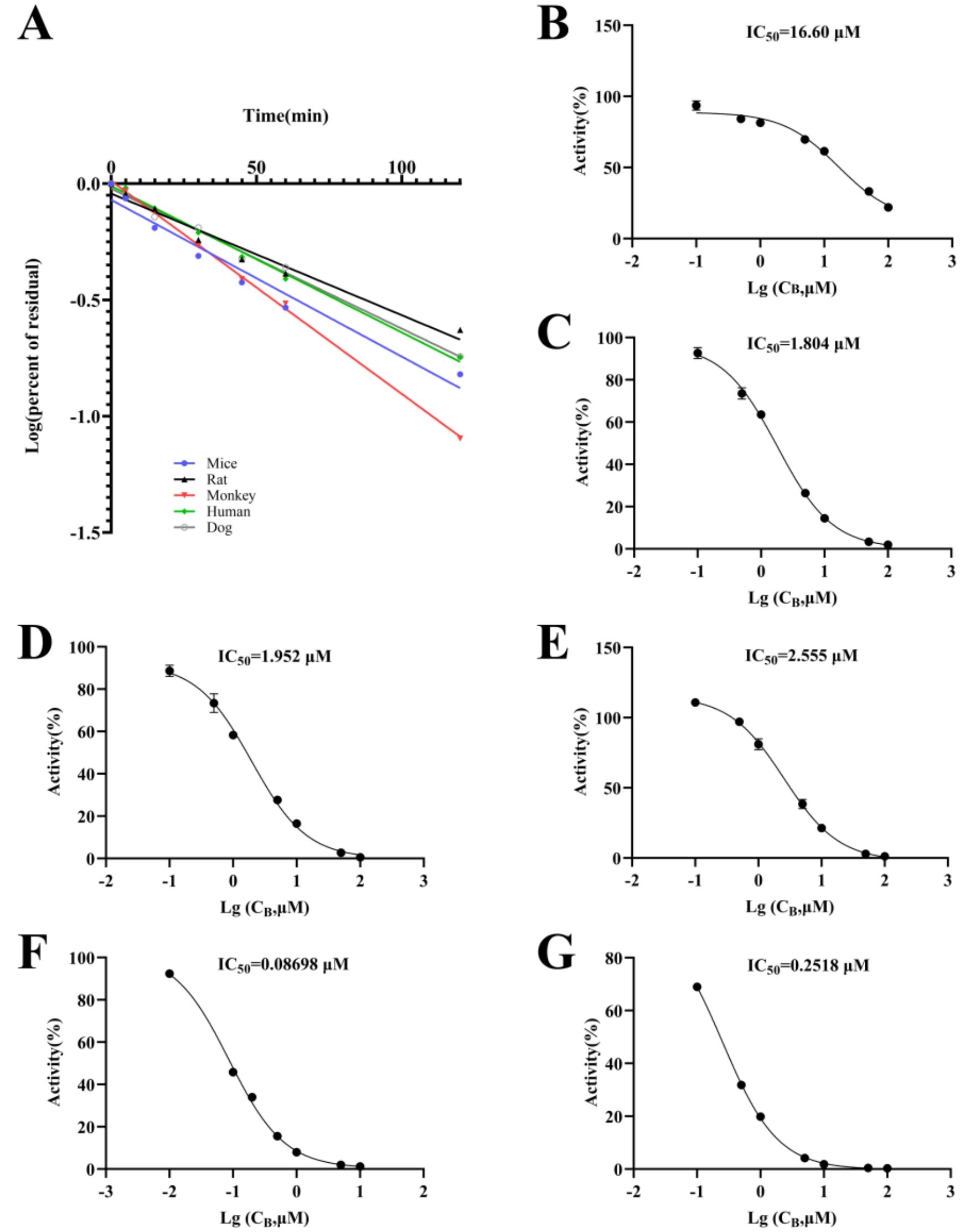

**Fig. 5 In vitro metabolic characterization and CYP450 inhibition profiles of Compound B. A.** Species-dependent metabolic stability of Compound B in liver microsomes from human, rat, mouse, dog, and monkey, highlighting interspecies variability in metabolic clearance. **B-G.** CYP450 inhibition profiles of Compound B across major drug-metabolizing enzymes. Panels B-G present probe substrate metabolism mediated by CYP1A2 (phenacetin), CYP2C9 (diclofenac), CYP2C19 (S-Mephenytoin), CYP2D6 (dextromethorphan), and CYP3A4 isoforms (testosterone and midazolam representing CYP3A4 C2 and M activity, respectively).

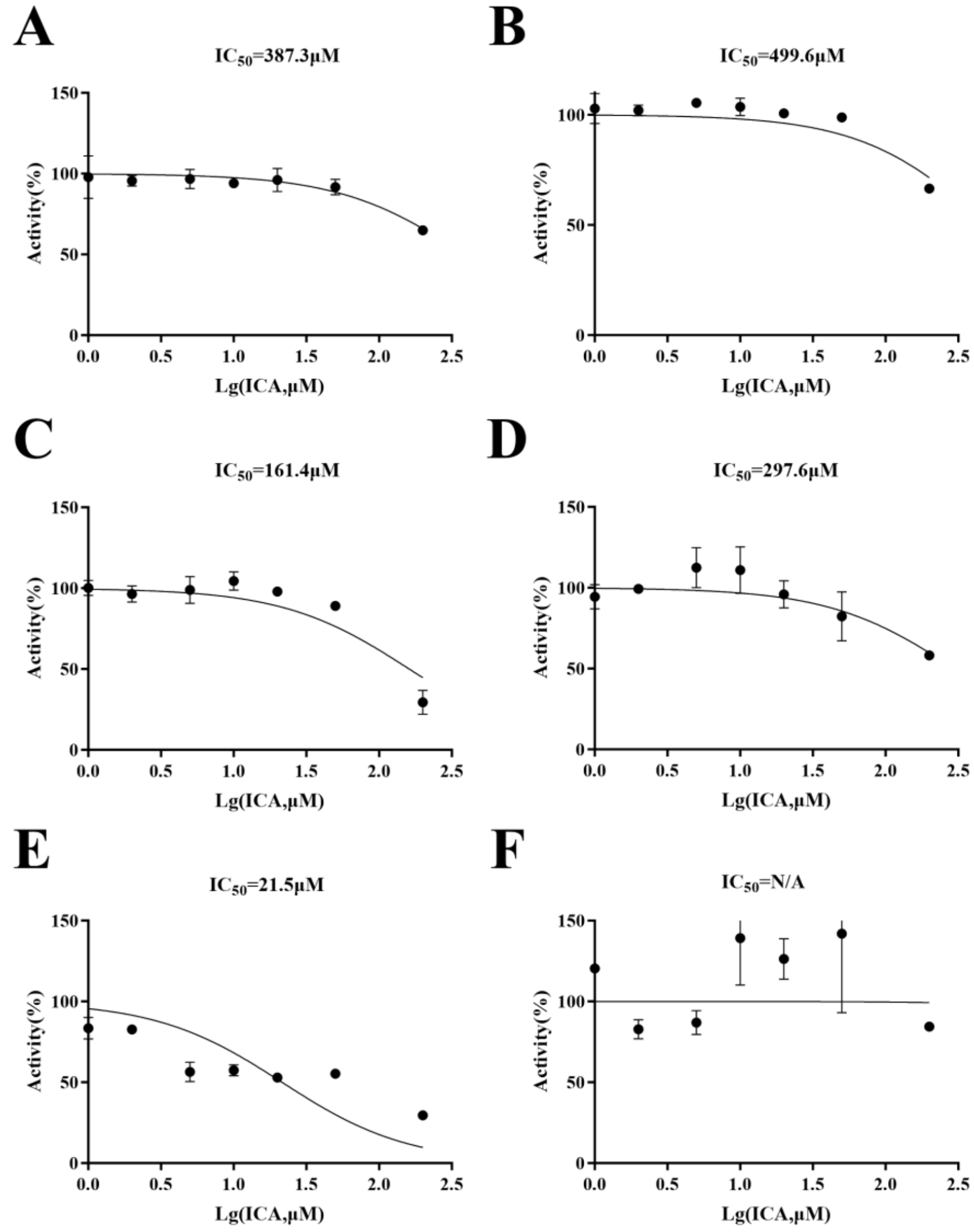

**Fig. 6 In vitro CYP450 inhibition profiles of Icariin. A-F.** In vitro CYP450 inhibition of Icariin across major drug-metabolizing enzymes. Panels A-F show probe substrate metabolism mediated by CYP1A2 (phenacetin), CYP2C9 (diclofenac), CYP2C19 (S-Mephenytoin), CYP2D6 (dextromethorphan), and CYP3A4 isoforms (testosterone and midazolam representing CYP3A4 C2 and M activity, respectively).

For Oroxylin A, the first-order rate constant (λ) was $6.6 \times 10^{-3}$ $min^{-1}$, corresponding to a half-life of 105.0 min, and a human liver microsomal clearance (HLMC) of 32.36 $\mu L \cdot min^{-1} \cdot mg\ protein^{-1}$ (Fig. 4A). Consistent with CYP450 profiling, Oroxylin A exhibited strong inhibition of CYP1A2 (IC50 = 0.5432 μM), moderate inhibition of CYP2C9 (IC50 = 11.97 μM, not meeting the inhibitor threshold of 10 μM used in this study), and weak inhibition toward other tested isoforms (IC50 > 44 μM) (Fig. 4B–G). Compound B displayed a similar metabolic clearance profile, with an HLMC of 31.5 $\mu L \cdot min^{-1} \cdot mg\ protein^{-1}$ (Fig. 5A), and showed broad CYP450 inhibitory activity across multiple enzymes except CYP1A2 (Fig. 5B–G). In contrast, Icariin exhibited high metabolic stability, with a markedly low clearance rate of 1 μL·min$^{-1}$·mg protein$^{-1}$, and showed no significant inhibition of the tested CYP450 enzymes (IC50 > 21.5 μM) (Fig. 6A–F). According to the annotation standard of the source dataset (PubChem AID: 1851), an IC50 threshold of 10 μM was used to define CYP450 inhibitors[47,48] in this study.

The corresponding properties were predicted using the best-performing MEGA-CL model selected from three independent training runs. As summarized in Table 3, for Oroxylin A, MEGA-CL correctly identified CYP1A2 inhibition and non-inhibition for CYP2C19, CYP2D6, and CYP3A4, with an HLMC relative error of 1.48 (47.93 vs 32.36 μL·min$^{-1}$·mg protein$^{-1}$). For Compound B, the model correctly classified four of five CYP450 endpoints, except CYP2C9, and achieved an HLMC relative error of 1.66(52.48 vs 31.50 μL·min$^{-1}$·mg protein$^{-1}$). For Icariin, three of five CYP450 endpoints were correctly predicted, with misclassifications observed for CYP2C9 and CYP2C19, and an HLMC relative error of 2.25 (2.25 vs 1.0 μL·min$^{-1}$·mg protein$^{-1}$). Overall, MEGA-CL achieved 73.3% accuracy across binary CYP450 classification tasks, and all HLMC predictions fell within a 2.5-fold error range.

**Table 3 External validation of MEGA-CL on Oroxylin A, Compoud B and Icariin.** This table summarizes computational predictions and corresponding experimental measurements for independent compounds selected for prospective validation of MEGA-CL.

| Property | Oroxylin A (Pred.) | Oroxylin A (Exp.) | Compoud B (Pred.) | Compoud B (Exp.) | Icariin (Pred.) | Icariin (Exp.) |
|---|---|---|---|---|---|---|
| HLMC | 47.9291 | 32.3593 | 52.4807 | 30.1500 | 2.2511 | 1.0000 |
| CYP1A2 inhibition | 0.7257 (Positive) | **Positive** | 0.3282 (Negative) | **Negative** | 0.2087 (Negative) | **Negative** |
| CYP2C9 inhibition | 0.8335 (Positive) | Negative | 0.3203 (Negative) | Positive | 0.9994 (Positive) | Negative |
| CYP2C19 inhibition | 0.0029 (Negative) | **Negative** | 0.7066 (Positive) | **Positive** | 0.9981 (Positive) | Negative |
| CYP2D6 inhibition | 0.0741 (Negative) | **Negative** | 0.6780 (Positive) | **Positive** | 0.1292 (Negative) | **Negative** |
| CYP3A4 inhibition | 0.4276 (Negative) | **Negative** | 0.9148 (Positive) | **Positive** | 0.2413 (Negative) | **Negative** |

The unit of HLMC is μL·min$^{-1}$·mg protein$^{-1}$. For the classification task (CYP450 inhibitors), the classification threshold was set to 0.5, correct predictions are in bold.

## External validation on clinically approved drugs

We also performed retrospective analysis using recently approved FDA drugs with publicly available in vitro metabolic data[49]. We curated recently approved FDA drugs with experimentally measured in vitro metabolic data, focusing on hepatic microsomal metabolic stability as quantified by intrinsic clearance (CLint), a widely used surrogate for in vivo drug disposition and metabolic liability[50,51]. In addition, we collected 2,284 human liver microsomal clearance (HLMC) records from the Pharmabench dataset[52] to ensure a broad coverage of chemical space and assay variability. The training results of HLMC and the other Pharmabench benchmark datasets are shown in Supplementary Table S7.

We curated 18 additional in vitro metabolic datasets from publicly available FDA documentation for drugs approved or newly indicated within the past five years. These compounds were selected based on the availability of complete HLMC measurements, covering a broad range of metabolic stability with intrinsic clearance (CLint) values spanning 10–60 $\mu L \cdot min^{-1} \cdot mg^{-1}$ protein[53,54]. The observed values, predicted values, fold errors, and calculated chemical properties of all predicted drugs, including relative molecular mass and logP, are shown in Supplementary Table S8. For comparison, RF, SVM, and XGBoost were selected as baseline models due to their strong performance in prior ADMET prediction benchmarks. Across these models, RF exhibited the largest number of high-deviation cases, with 5 drugs exceeding 5-fold error and only 7 within the 2-fold range. SVM showed improved stability, with only 1 drug exceeding 5-fold error and 4 exceeding 2-fold deviation, while XGBoost demonstrated intermediate performance with 2 drugs exceeding 5-fold error and moderate distribution across 2–3-fold error ranges (Fig. 7A–C). In contrast, MEGA-CL achieved consistently higher accuracy, with 16 of 18 drugs predicted within a 2-fold error range and all remaining compounds within 3-fold deviation (Fig. 7D). Collectively, these results demonstrate that MEGA-CL maintains stable predictive performance across clinically relevant external drug datasets and exhibits improved robustness compared with conventional machine learning baselines.

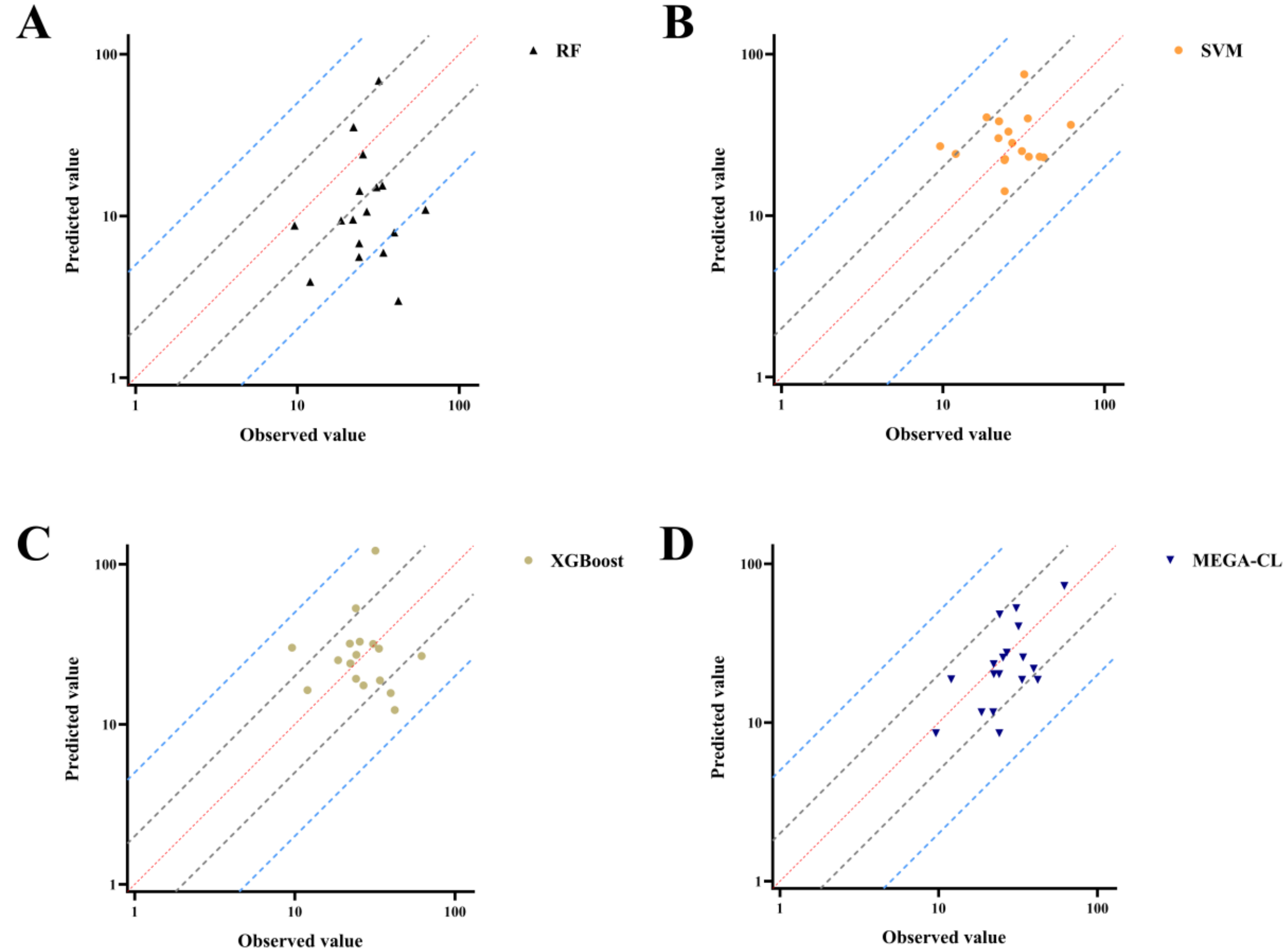

**Fig. 7 External validation of HLMC prediction on 21 independent drugs.** Observed versus predicted human liver microsomal clearance (HLMC) values for 21 independent compounds are shown on a logarithmic scale. Panels A-D correspond to Random Forest (RF), Support Vector Machine (SVM), XGBoost, and MEGA-CL, respectively. The red diagonal indicates perfect prediction, while gray and blue bands represent 2-fold and 5-fold error ranges, respectively, used to assess prediction accuracy and deviation magnitude.

## Conclusions

In this study, we developed MEGA-CL, a graph external attention–enhanced contrastive learning framework for the prediction of ADMET properties, with a particular focus on hepatic metabolic clearance and CYP450 inhibition. The model integrates an enhanced GCN+ backbone with residual connections and layer normalization, together with a multi-head graph external attention mechanism. This design effectively alleviates the over-smoothing limitation of conventional graph neural networks and enables simultaneous learning of local molecular substructures and global inter-graph chemical space relationships, thereby producing more expressive molecular representations.

MEGA-CL adopts a two-stage pre-training and fine-tuning paradigm that leverages large-scale unlabeled molecular data for self-supervised representation learning. Pre-training on approximately 100 million molecules allows the model to capture generalizable chemical knowledge, which is then efficiently transferred to downstream ADMET prediction tasks through task-specific fine-tuning. This strategy significantly improves model generalization, as demonstrated by consistent performance gains across both benchmark datasets and small-scale tasks.

Extensive evaluations on MoleculeNet benchmarks and 21 ADMET prediction tasks demonstrate that MEGA-CL consistently outperforms traditional machine learning methods as well as strong deep learning baselines. In particular, the model shows strong performance in both classification and regression settings, achieving superior accuracy in challenging endpoints such as intrinsic clearance and volume of distribution. Fold-error analysis further confirms its practical utility, with the majority of predictions falling within acceptable error ranges and reduced systematic bias compared with baseline models.

To evaluate real-world applicability, we conducted comprehensive external validation using 18 FDA-approved or newly indicated drugs and three structurally diverse preclinical compounds, forming a 21-molecule independent test set. Across a wide range of intrinsic clearance values, MEGA-CL maintained robust predictive performance, with most compounds predicted within a 2-fold error range. The model also achieved strong CYP450 classification performance and demonstrated good generalization to unseen chemical scaffolds, while baseline models exhibited greater variability and higher error rates.

In conclusion, MEGA-CL provides a robust and scalable computational framework for ADMET prediction. By integrating self-supervised contrastive learning with graph external attention, it enables accurate and generalizable prediction of drug metabolic properties, offering a practical tool to accelerate early-stage drug discovery and lead optimization.

## Methods

### Software and environment

The computational tasks in this study were implemented based on the Python 3.7.12 environment. TensorFlow 2.11.0 and PyTorch 1.13.1+cu11.6 were adopted as the deep learning tools to construct the neural network models, while PyTorch Geometric 2.0.1 provided core support for the implementation of graph neural networks (GNNs)[55–57]. Data processing and scientific computing relied on SciPy 1.7.3 and Scikit-learn 1.0.2. RDKit-PyPI 2022.9.5 was employed for molecular structure processing, including parsing compound structures from SMILES strings, constructing molecular graphs, and extracting chemical features[58].

### Dataset compilation and preprocessing

To enable robust self-supervised representation learning, we constructed a massive pre-training dataset by querying the PubChem database. From an initial retrieval of approximately $1.7 \times 10^{8}$ molecular data entries, rigorous filtering was applied to remove duplicates, invalid structures, and disconnected fragments (e.g., salts and solvents). This yielded approximately 10 million unique, unannotated molecular SMILES strings. For downstream evaluation, we utilized 13 benchmark datasets from MoleculeNet, encompassing diverse physiological, biophysical, and quantum mechanical properties[35]. Similarly, we obtained nine tasks from the large-scale LLM-based Pharmabench dataset, and the corresponding training results are provided in the Supplementary Materials[52]. Furthermore, we curated an extensive ADMET-specific dataset comprising 19 tasks (including highly complex endpoints such as steady-state volume of distribution(VDss), and clearance,(CL), meticulously standardized using the RDKit chemoinformatics software suite[59,60]. Scaffold splitting with an 8:1:1 ratio (train/validation/test) was consistently applied to simulate real-world prospective drug discovery scenarios.

This study systematically constructed a high-quality ADMET dataset through public data retrieval, literature collation, and standardized manual curation to support comprehensive drug ADMET modeling. In this work, we prioritized sample adequacy (more than 1,000 valid entries per task) and physiological relevance for dataset construction and preferentially adopted human experimental data to guarantee translational value. All datasets were strictly evaluated throughout model training to ensure reliable modeling performance, regardless of ideal statistical data distribution. For ADMET classification tasks, the core datasets were primarily curated from PubChem BioAssay (AID: 1851)[47]. A total of 16,560 raw records were initially collected and further standardized via SMILES normalization and binary label transformation. With a unified $IC_{50}$ threshold of 10 μM, we established classification datasets to evaluate the inhibitory activities of five major human hepatic CYP isoforms, including CYP1A2, CYP2C9, CYP2D6, CYP2C19, and CYP3A4. Following the same processing pipeline, we supplemented P-gp substrate classification data and PAMPA permeability classification data from public resources and published literature to support the assessment of drug absorption properties[61]. We also incorporated an auxiliary CYP enzyme dataset containing approximately 2,000 molecules per isoform for model training from the study by Yu-Hao Ni et al[60]. Considering the insufficient quantity, uneven distribution, and poor availability of valid assay data for minor hepatic enzymes (e.g., CYP2B6 and CYP2C8), these isoforms were excluded from the current study to ensure overall data quality. For complex pharmacokinetic property prediction, we retrieved a total of 1,352 clinically derived drug data points covering five core in vivo PK indices: Fu, VDss, CL, MRT, and HL from the study by Franco Lombardo et al[59]. Despite the presence of outliers, partial missing values, and multi-source heterogeneity in this dataset, it provides valuable real in vivo pharmacological measurements and was retained as the fundamental data resource for PK regression modeling. Toxicity modeling suffers from limited standardized datasets and ambiguous structure–toxicity relationships. We constructed two refined toxicity datasets via standardized curation strategies: the Ames mutagenicity dataset was obtained from PharmaBench[52] through an LLM-assisted data cleaning pipeline, and a binary drug-induced liver injury dataset (DIList) was built by extracting and relabeling high-risk and low-risk samples from FDA clinical reports[62].

After rigorous data collection, collation, relabeling, and quality control, we finally established a comprehensive ADMET dataset system covering 12 classification tasks and 7 regression tasks across the entire drug absorption, distribution, metabolism, and excretion process. All molecular structures were verified against PubChem and standardized with unified SMILES strings, and all datasets were formatted with consistent headers for model fine-tuning. Detailed data annotation and preprocessing protocols are summarized in Table S1.

## Molecular graph construction and augmentations

Let a molecule be represented as an undirected graph $\mathcal{G} = (\mathcal{V}, \mathcal{E})$, where $\mathcal{V}$ enotes the set of atoms (nodes) and $\mathcal{E}$ denotes the set of chemical bonds (edges). The initial node features $\mathbf{h}_v^{(0)} \in \mathbb{R}^{d_{in}}$ encompass atomic number, chirality, degree, formal charge, and hybridization. Following the MolCLR framework, we apply three stochastic graph augmentation strategies during pre-training to generate correlated views: (1) Atom masking, where node features are randomly masked with a probability $p_m$ (2) Bond deletion, which drops chemical bonds and their corresponding edge features with a probability $p_b$; and (3) Subgraph removal, which simulates missing functional groups by removing random subgraphs using a random walk approach.

## Enhanced message-passing architecture (GCN+)

To overcome the inherent limitations of classic message-passing neural networks—namely, over-smoothing and vanishing gradients—we adopted the GCN+ architecture as our intra-graph feature extractor. The node update rule at the $l$th layer integrates neighborhood aggregation, batch normalization (BN), and non-linear activation (ReLU), formulated as

$$\boldsymbol{z}_v^{(l)} = \mathrm{ReLU}\left(\mathrm{BN}\left(\sum_{u\in\mathcal{N}(v)\cup\{v\}} \frac{1}{\sqrt{\tilde{d}_u\tilde{d}_v}} \boldsymbol{W}^{(l)}\boldsymbol{h}_u^{(l-1)}\right)\right) \tag{1}$$

is the neighborhood of node $v$, $\tilde{d}_v$is the node degree incorporating self-loops, and $\mathbf{W}^{(l)}$is a learnable weight matrix. Crucially, GCN+ incorporates identity residual connections and a Feed-Forward Network (FFN) augmented with Dropout to preserve deep structural integrity

$$\boldsymbol{h}_v^{(l)} = \boldsymbol{h}_v^{(l-1)} + \mathrm{Dropout}\left(\mathrm{FFN}(\boldsymbol{z}_v^{(l)})\right) \tag{2}$$

The global representation of the graph, $\mathbf{h}_{\mathcal{G}} \in \mathbb{R}^{d_{model}}$ is aggregated via a mean-pooling readout function

$$\boldsymbol{h}_{\mathcal{G}} = \frac{1}{|\mathcal{V}|} \sum_{v \in \mathcal{V}} \boldsymbol{h}_v^{(L)} \quad (3)$$

**Multi-Head Graph External Attention (GEA)**

Traditional graph attention mechanisms are confined to intra-graph interactions. To implicitly capture inter-graph chemical correlations, we integrate a Multi-Head Graph External Attention (GEA) module. The GEA introduces two learnable external memory matrices: the Key matrix $\mathbf{M}_K \in \mathbb{R}^{S \times d_k}$ where $S$ is the memory size representing latent global graph prototypes. Given the intra-graph representation $\mathbf{h}_{\mathcal{G}}$ the query vector is computed as $\mathbf{Q} = \mathbf{h}_{\mathcal{G}} \mathbf{W}_Q$ The external attention output $\mathbf{E} \in \mathbb{R}^{d_v}$ for a single head is computed via scaled dot-product attention against the global memory: $\mathbf{E} = \mathrm{Softmax}\left(\frac{\mathbf{Q}\mathbf{M}_K^{\top}}{\sqrt{d_k}}\right)\mathbf{M}_V$ For the multi-head mechanism with $H$ heads, the outputs are concatenated and linearly projected. The final graph representation $\mathbf{Z}_{\mathcal{G}}$ fuses local structure and global interaction contexts via layer normalization (LN)

$$\boldsymbol{Z}_{\mathcal{G}} = \mathrm{LN}\big(\boldsymbol{h}_{\mathcal{G}} + \mathrm{Concat}(\boldsymbol{E}_1, \dots, \boldsymbol{E}_H)\boldsymbol{W}_O\big) \quad (4)$$

**Self-supervised contrastive learning protocol**

During pre-training, each molecule $\mathcal{G}$ in a minibatch of size N = 512 is subjected to two independent random augmentations, resulting in 2N augmented graphs. The model is trained using the Normalized Temperature-scaled Cross-Entropy (NT-Xent) loss to maximize the agreement between latent representations $\mathbf{Z}_i \ and\ \mathbf{Z}_j$ of the same molecule, while repelling representations of different molecules. The contrastive loss for a positive pair $(i, j)$ is defined as

$$\ell_{i,j} = -log \frac{exp\ (\mathrm{sim}(\boldsymbol{Z}_i, \boldsymbol{Z}_j)/\tau)}{\sum_{k=1}^{2N} \mathbb{1}_{[k \neq i]} exp\ (\mathrm{sim}(\boldsymbol{Z}_i, \boldsymbol{Z}_k)/\tau)} \quad (5)$$

where

$$\mathrm{sim}(\boldsymbol{u}, \boldsymbol{v}) = \frac{\boldsymbol{u}^{\top}\boldsymbol{v}}{\|\boldsymbol{u}\|\|\boldsymbol{v}\|} \quad (6)$$

denotes the cosine similarity, $\tau$ is the temperature hyperparameter, and $\mathbb{1}_{[k \neq i]}$ is an indicator function evaluating to 1 if $k \neq i$ . Pre-training was executed for 100 epochs using the Adam optimizer with a weight decay of $10^{-5}$ and an initial learning rate of $5 \times 10^{-4}$. A learning rate warm-up strategy was implemented, followed by cosine decay.

**Fine-tuning and evaluation metrics**

For downstream tasks, the pre-trained encoder weights were transferred, and a task-specific MLP prediction head was randomly initialized. Fine-tuning was conducted with a batch size of 32 for 100 epochs. Classification tasks were optimized using cross-entropy loss and evaluated via the Area Under the Receiver Operating Characteristic Curve (ROC-AUC). Regression tasks were optimized via $\ell_1$ loss (Mean Absolute Error) and evaluated using Root Mean Square Error (RMSE), Mean Absolute Error (MAE), and the coefficient of determination($R^2$).

$$RMSE = \sqrt{\sum_{i=1}^{n} \frac{(\hat{y}_i - y_i)^2}{n}} \tag{7}$$

$$MAE = \sum_{i=1}^{n} \frac{(\hat{y}_i - y_i)^2}{n} \tag{8}$$

$$R^2 = 1 - \frac{\sum_i (\hat{y}_i - y_i)^2}{\sum_i (\overline{y}_i - y_i)^2} \tag{9}$$

represent the experimental value, predicted value, and the mean of experimental values, respectively. For baseline comparisons, five machine learning methods (Random Forest, SVM, XGBoost, Logistic Regression/Classification, and Naive Bayes) were employed alongside the ChemProp deep learning baseline.To rigorously assess the translational validity of ADMETT predictions, we additionally utilized the Average Fold Error, an essential metric in pharmacokinetics defined as

$$Fold\ Error(i) = max(\frac{\hat{y}_i}{y_i}, \frac{y_i}{\hat{y}_i}) \tag{10}$$

Robustness was evaluated by quantifying the percentage of molecules falling within 2-fold, 3-fold, and 5-fold error margins. All downstream fine-tuning experiments were repeated independently three times, with the mean and standard deviation reported to ensure statistical reliability.

## Baseline

For classification tasks, five machine learning methods were employed: RF, SVM, XGBoost, Logistic Classification, and Naive Bayes[42–44,63]. For regression tasks, we used RF, SVM, XGBoost, and Logistic Regression. Meanwhile, we constructed the baseline for comparison in this study by referring to Chemprop, a molecular property learning method based on DMPNN[46].

## Incubation operation for in vitro metabolic stability assay

The *in vitro* phase I metabolic incubation mixture contained $MgCl_2$ (5.0 mM), NADPH (1.0 mM), PBS buffer (pH 7.2–7.4, 0.1M), test compound working solution and liver microsomes (human, rat, mouse, monkey, dog) ( 0.2 mg protein/ml). The final concentration of each compound in incubation system was set as follows: Oroxylin A at 10 μM, compound B at 0.5 μM, Icariin at 5 μM.Test compounds and liver microsomes were pre-incubated at 37 °C for 5 min; reactions were initiated by NADPH addition, with sampling time points set at 0, 2, 5, 15, 30, 45, 60, 120 min. The organic solvent fraction in the system was controlled ≤1%, and blank controls without test compounds were incubated for 120 min in parallel. Post-incubation, ice-cold internal standard solution was immediately added to quench reactions. Samples were vortexed and centrifuged, and supernatants were collected for UPLC-MS/MS detection.

Metabolic constants and clearance rates were calculated using the following formulas: The apparent first-order rate constant (k) was calculated by linear regression of ln (remaining substrate concentration) versus incubation time, with the formula: $\ln(C_t) = -kt + \ln(C_0)$ where $C_t$ is the concentration of substrate at time t, $C_0$ is the initial concentration at time 0, and t is the incubation time. The half-life ($t_{1/2}$) was calculated as $t_{1/2} = \frac{\ln 2}{k}$ The intrinsic clearance ($CL_{int}$) was calculated using the formula: $CL_{int} = \frac{k \times V}{C_{protein}}$ where V is the incubation volume, and the microsomal protein concentration is 0.2 mg/mL in this assay.

The incubation system for CYP450 reversible inhibition assay contained $MgCl_2$ (5.0 mM), NADPH (1.0 mM), PBS buffer (pH 7.2–7.4, 0.1M), and human liver microsomes (0.2 mg protein/ml), serial concentrations of test compound, subtype-specific probe substrates, 0.2 mg/mL human liver microsomes and 1 mM NADPH initiator. The organic solvent proportion was less than 1%. The probe substrates matched with each CYP isoform and their fixed final concentrations were listed below: CYP1A2: phenacetin 60 μM; CYP2B6: bupropion 140 μM; CYP2C8: amodiaquine 10 μM; CYP2C9: diclofenac 6 μM; CYP2C19: S-mephenytoin 20 μM; CYP2D6: dextromethorphan 5 μM; CYP3A4(C2): testosterone 50 μM; CYP3A4(M): midazolam 10 μM. Each isoform adopted matched incubation time: CYP1A2 for 20 min, CYP2B6 and CYP2C19 for 30 min, CYP2C8 and CYP3A4(C2) for 10 min, CYP2C9 and CYP2D6 for 15 min, CYP3A4(M) for 5 min. After incubation, ice internal standard solution was added to terminate reaction, and samples were processed for UPLC-MS/MS detection. Serial concentration gradients of three test compounds were prepared as follows: (1) Oroxylin A: 0.1, 0.2, 0.5, 1, 2, 5, 10, 30 μM; (2) Compound B: 0.1, 0.5, 1, 5, 10, 50, 100 μM for most CYP isoforms; 0.1, 0.5, 1, 2, 5, 10, 15 μM for CYP2C19; 0.01, 0.1, 0.2, 0.5, 1, 5, 10 μM for CYP3A4(C2); (3) Icariin: 0.1, 0.2, 0.5, 1, 2, 5, 10, 20, 50 μM (50 μM was the solubility upper limit).

The inhibition rate of each test compound at different concentrations was calculated, then inhibition rate was plotted against the logarithmic concentration of test compounds. Half maximal inhibitory concentration ($IC_{50}$) values were calculated via nonlinear regression fitting using GraphPad Prism software. The standard operating procedure for phase I microsomal metabolic assays and CYP450 inhibition tests was documented in the supplementary protocol, while the incubation system composition, probe substrate incubation parameters, and mass spectrometric MRM transition settings were separately summarized in Table S9-S11, respectively.

## Data Availability

The pre-training data and molecular property prediction benchmarks used in this work are available in the Zenodo https://doi.org/10.5281/zenodo.20793660.

## Code Availability

All code was implemented in Python. Neural networks were implemented using Pytorch. Both the code and scripts to reproduce the experiments of this paper are available at https://github.com/KeduJin/MEGA-CL.

## Author contributions

J.L., D.Z., and X.C. conceived and supervised the research and designed the experiments. T.J., K.J., Y.L., G.R., J.X., S.Z., X.D., and L.W. established, validated the MEGA-CL model and analyzed the data. T.J. and K.J. wrote the manuscript. J.L., D.Z., and X.C. revised the manuscript.

## Acknowledgements

This work was supported by the National Natural Science Foundation of China (Grant 22373116), and Jiangsu Key Research and Development Program (Grant BE2022365). The Supercomputer Center of China Pharmaceutical University is acknowledged for providing computer resources.

## Competing interests

The authors declare no competing interests.

## Supplementary Information

The standard operating procedure for in vitro hepatic microsomal metabolic and CYP450 inhibition assays and Supplementary Tables 1–11, Supplementary Figs. 1–3.

## References

1. Hughes, J., Rees, S., Kalindjian, S. & Philpott, K. Principles of early drug discovery. *Br. J. Pharmacol.* **162**, 1239–1249 (2011).

2. Zhang, K. *et al.* Artificial intelligence in drug development. *Nat. Med.* **31**, 45–59 (2025).

3. Yu, L. X. Pharmaceutical Quality by Design: Product and Process Development, Understanding, and Control. *Pharm. Res.* **25**, 781–791 (2008).

4. Waring, M. J. *et al.* An analysis of the attrition of drug candidates from four major pharmaceut ical companies. *Nat. Rev. Drug Discovery* **14**, 475–486 (2015).

5. Sun, D., Gao, W., Hu, H. & Zhou, S. Why 90% of clinical drug development fails and how to improve it? *Acta Pharm. Sin., B* **12**, 3049–3062 (2022).

6. Kerns, E. H. & Di, L. Drug-like Properties: Concepts, Structure Design and Methods: From AD ME to Toxicity Optimization. (Academic Press, Amsterdam ; Boston, 2008).

7. Ito, K. & Houston, J. B. Comparison of the Use of Liver Models for Predicting Drug Clearanc e Using in Vitro Kinetic Data from Hepatic Microsomes and Isolated Hepatocytes. *Pharm. Res.* **21**, 785–792 (2004).

8. Qiang, B. *et al.* Bridging the gap between chemical reaction pretraining and conditional molecul e generation with a unified model. *Nat Mach Intell* **5**, 1476–1485 (2023).

9. Wang, W. *et al.* An Integrated AI-PBPK Platform for Predicting Drug In Vivo Fate and Tissue Distribution in Human and Inter-Species Extrapolation. *Clin. Pharmacol. Ther.* **118**, 865–875 (2 025).

10. Sliwoski, G., Kothiwale, S., Meiler, J. & Lowe, E. W. Computational Methods in Drug Discove ry. *Pharmacol. Rev.* **66**, 334–395 (2014).

11. Sadybekov, A. V. & Katritch, V. Computational approaches streamlining drug discovery. *Nature* **616**, 673–685 (2023).

12. Venkataraman, M., Rao, G., Reddi, J. & Maddi, S. Leveraging machine learning models in eval uating ADMET properties for drug discovery and development. *ADMET DMPK* **13**, 2772 (202 5).

13. Suvidhi, M., Kumar, S., Sumanshu, M. & Vaid, R. In silico admet predictions: enhancing drug development through qsar modeling. in 41–52 (2024). doi:10.58532/V3BCAG21P2CH1.

14. Gilmer, J., Schoenholz, S. S., Riley, P. F., Vinyals, O. & Dahl, G. E. Neural Message Passing for Quantum Chemistry. Preprint at https://arxiv.org/abs/1704.01212v2 (2017).

15. Schütt, K. T. *et al.* SchNet: A continuous-filter convolutional neural network for modeling quan tum interactions. Preprint at https://arxiv.org/abs/1706.08566v5 (2017).

16. Guo, Z. *et al.* Graph-based molecular representation learning. Preprint at https://doi.org/10.48550 /arXiv.2207.04869 (2023).

17. Dinh Pham, L.-H., Le, M.-T. & Thai, K.-M. Improved ADME Prediction by Multitask Pretraini ng on Predicted Data: Insights from the ASAP-Polaris-OpenADMET Blind Challenge. *J. Chem. Inf. Model.* **66**, 395–405 (2026).

18. Putrama, I. M. & Martinek, P. Heterogeneous data integration: Challenges and opportunities. *Da ta Brief* **56**, 110853 (2024).

19. Pallikkavaliyaveetil, N. & Chandrasekaran, S. Small data, big challenges: Machine- and deep-lea rning strategies for data-limited drug discovery. *Adv. Drug Delivery Rev.* **229**, 115762 (2026).

20. Liu, Y., Lim, H. & Xie, L. Exploration of chemical space with partial labeled noisy student sel f-training and self-supervised graph embedding. *BMC Bioinf.* **23**, 158 (2022).

21. Wang, Y., Wang, J., Cao, Z. & Barati Farimani, A. Molecular contrastive learning of representa tions via graph neural networks. *Nat. Mach. Intell.* **4**, 279–287 (2022).

22. You, Y. *et al.* Graph Contrastive Learning with Augmentations. Preprint at https://arxiv.org/abs/2010.13902 (2020).

23. Chen, Y. *et al.* Self-supervised learning in drug discovery. *Sci. China Inf. Sci.* **68**, 170103 (202 5).

24. Xia, J., Zhu, Y., Du, Y. & Li, S. Z. Pre-training Graph Neural Networks for Molecular Represe ntations: Retrospect and Prospect. in *ICML 2022 2nd AI for Science Workshop* (2022).

25. Li, X. *et al.* Neural atoms: propagating long-range interaction in molecular graphs through effici ent communication channel. Preprint at https://doi.org/10.48550/arXiv.2311.01276 (2024).

26. Godwin, J. *et al.* Simple GNN regularisation for 3D molecular property prediction & beyond. P reprint at https://doi.org/10.48550/arXiv.2106.07971 (2022).

27. Zhao, B., Xu, W., Guan, J. & Zhou, S. Molecular property prediction based on graph structure learning. *Bioinformatics* **40**, btae304 (2024).

28. Klopmand, G. Concepts and applications of molecular similarity, by Mark A. Johnson and Gera ld M. Maggiora, eds., John Wiley & Sons, New York, 1990, 393 pp. Price: $65.00. *J. Comput. Chem.* **13**, 539–540 (1992).

29. Chithrananda, S., Grand, G. & Ramsundar, B. ChemBERTa: Large-Scale Self-Supervised Pretrai ning for Molecular Property Prediction. Preprint at https://api.semanticscholar.org/CorpusID:224803102 (2020).

30. Li, G., Müller, M., Thabet, A. & Ghanem, B. DeepGCNs: can GCNs go as deep as CNNs? Pre print at https://doi.org/10.48550/arXiv.1904.03751 (2019).

31. Li, Y., Wan, Y. & Liu, X. Semi-supervised Learning with Graph Convolutional Networks Base d on Hypergraph. *Neural Process. Lett.* **54**, 2629–2644 (2022).

32. Guo, M.-H., Liu, Z.-N., Mu, T.-J. & Hu, S.-M. Beyond Self-Attention: External Attention Usin g Two Linear Layers for Visual Tasks. *IEEE Trans. Pattern Anal. Mach. Intell.* **45**, 5436–5447 (2023).

33. Méndez-Lucio, O., Nicolaou, C. A. & Earnshaw, B. MolE: a foundation model for molecular gr aphs using disentangled attention. *Nat. Commun.* **15**, 9431 (2024).

34. Liu, J. *et al.* Towards Graph Foundation Models: A Survey and Beyond. Preprint at https://doi.org/10.48550/arXiv.2310.11829 (2023).

35. Wu, Z. *et al.* MoleculeNet: A Benchmark for Molecular Machine Learning. Preprint at https://arxiv.org/abs/1703.00564 (2018).

36. Lu, C. *et al.* Molecular Property Prediction: A Multilevel Quantum Interactions Modeling Persp ective. Preprint at https://arxiv.org/abs/1906.11081 (2019).

37. Yang, K. *et al.* Analyzing Learned Molecular Representations for Property Prediction. *J. Chem. Inf. Model.* **59**, 3370–3388 (2019).

38. Errica, F., Podda, M., Bacciu, D. & Micheli, A. A Fair Comparison of Graph Neural Networks for Graph Classification. Preprint at https://arxiv.org/abs/1912.09893 (2022).

39. Liu, S., Demirel, M. F. & Liang, Y. N-gram graph: simple unsupervised representation for grap hs, with applications to molecules. Preprint at https://doi.org/10.48550/arXiv.1806.09206 (2019).

40. Mobley, D. L. & Guthrie, J. P. FreeSolv: a database of experimental and calculated hydration f ree energies, with input files. *J. Comput.-Aided Mol. Des.* **28**, 711–720 (2014).

41. Hu, W. *et al.* Open graph benchmark: datasets for machine learning on graphs. Preprint at http s://doi.org/10.48550/arXiv.2005.00687 (2021).

42. Breiman, L. Random Forests. *Mach. Learn.* **45**, 5–32 (2001).

43. Cortes, C. & Vapnik, V. Support-vector networks. *Mach. Learn.* **20**, 273–297 (1995).

44. Chen, T. & Guestrin, C. XGBoost: A Scalable Tree Boosting System. in *Proceedings of the 22 nd ACM SIGKDD International Conference on Knowledge Discovery and Data Mining* 785–794 (Association for Computing Machinery, New York, NY, USA, 2016). doi:10.1145/2939672.293 9785.

45. Berkson, J. Application of the logistic function to bio-assay. *J. Am. Stat. Assoc.* **39**, 357–365 (2 012).

46. Heid, E. *et al.* Chemprop: a machine learning package for chemical property prediction. *J. Che m. Inf. Model.* **64**, 9–17 (2024).

47. National Center for Biotechnology Information. PubChem BioAssay: Quantitative High-Through put Screening for Inhibitors of Human Cytochrome P450 Isoforms 1A2, 2C9, 2C19, 2D6, and 3 A4 (AID: 1851). (2009).

48. In Vitro Drug Interaction Studies — Cytochrome P450 Enzyme- and Transporter-Mediated Dru g Interactions Guidance for Industry. (U.S. Food and Drug Administration, 2020).

49. Attwa, M. W., AlRabiah, H., Mostafa, G. A. E., Bakheit, A. H. & Kadi, A. A. Assessment of I n Silico and In Vitro Selpercatinib Metabolic Stability in Human Liver Microsomes Using a Va lidated LC-MS/MS Method. *Molecules* **28**, 2618 (2023).

50. Sodhi, J. K. & Benet, L. Z. Successful and Unsuccessful Prediction of Human Hepatic Clearanc e for Lead Optimization. *J. Med. Chem.* **64**, 3546–3559 (2021).

51. Obach, R. S. *et al.* The Prediction of Human Pharmacokinetic Parameters from Preclinical and *In Vitro* Metabolism Data. *J. Pharmacol. Exp. Ther.* **283**, 46–58 (1997).

52. Niu, Z. *et al.* PharmaBench: Enhancing ADMET benchmarks with large language models. *Sci. Data* **11**, 985 (2024).

53. Shah, P., Siramshetty, V. B., Mathé, E. & Xu, X. Developing Robust Human Liver Microsomal Stability Prediction Models: Leveraging Inter-Species Correlation with Rat Data. *Pharmaceutics* **16**, 1257 (2024).

54. Ryu, J. Y. *et al.* PredMS: a random forest model for predicting metabolic stability of drug cand idates in human liver microsomes. *Bioinformatics* **38**, 364–368 (2022).

55. Abadi, M. *et al.* TensorFlow: Large-Scale Machine Learning on Heterogeneous Distributed Syst ems. Preprint at https://arxiv.org/abs/1603.04467 (2016).

56. Paszke, A. *et al.* PyTorch: An Imperative Style, High-Performance Deep Learning Library. Prep rint at https://arxiv.org/abs/1912.01703 (2019).

57. Fey, M. & Lenssen, J. E. Fast Graph Representation Learning with PyTorch Geometric. Preprin t at https://arxiv.org/abs/1903.02428 (2019).

58. Pedregosa, F. Scikit-learn: Machine Learning in Python. *J. Mach. Learn. Res.* **12**, 2825–2830 (2 011).

59. Lombardo, F., Berellini, G. & Obach, R. S. Trend Analysis of a Database of Intravenous Phar macokinetic Parameters in Humans for 1352 Drug Compounds. *Drug Metab. Dispos.* **46**, 1466–1477 (2018).

60. Fang, J. *et al.* Prediction of Cytochrome P450 Substrates Using the Explainable Multitask Deep Learning Models. *Chem. Res. Toxicol.* **37**, 1535–1548 (2024).

61. Parasar, U., Baruah, O., Saikia, D., Bharali, P. & Mahanta, H. J. Interpretable machine learning and graph attention network based model for predicting PAMPA permeability. *J. Mol. Graphics Model.* **138**, 109050 (2025).

62. Thakkar, S. *et al.* Drug-induced liver injury severity and toxicity (DILIst): binary classification of 1279 drugs by human hepatotoxicity. *Drug Discovery Today* **25**, 201–208 (2020).

63. Wickramasinghe, I. & Kalutarage, H. Naive bayes: applications, variations and vulnerabilities: a review of literature with code snippets for implementation. *Soft Comput.* **25**, 2277–2293 (2021).

# Supplementary Material for

# MEGA-CL: A Molecular Foundation Model for Generalizable ADMET Prediction through Graph External Attention and Contrastive Learning

*Tinghui Jin[1,#], Kedu Jin[1,#], Ying Li[1], Guanghui Ren[1], Jingzhi Xue[1], Shiyu Zhou[1], Xiaoli Dai[2], Libin Wei[3], Xijing Chen[1,*], Di Zhao[1,*], Jinfeng Liu[1,*]*

*[1]School of Science, School of Basic Medicine and Clinical Pharmacy, China Pharmaceutical University, Nanjing, China*

*[2]Sir Run Run Hospital, Nanjing Medical University, Nanjing, China*

*[3]Jiangsu Key Laboratory of Carcinogenesis and Intervention, China Pharmaceutical University, Nanjing, China*

*To whom correspondence should be addressed: jinfengliu@cpu.edu.cn (J.L.), zh_d99@cpu.edu.cn (D.Z.), chenxj-lab@hotmail.com (X.C,)

[#]These authors contributed equally to this work

---

## Standard Operating Procedure for *In Vitro* Hepatic Microsomal Metabolism and CYP450 Inhibition Assays

## 1. Reagents, Chemicals and Hepatic Microsome Source

### 1.1 Test Articles

Oroxylin A (purity, 99.20%; batch number: 20171106) was purchased from China Pharmaceutical University. This compound was hermetically sealed, stored away from light at 2–8 °C, and had a shelf life of 2 years. Its stock solution was prepared in dimethyl sulfoxide (DMSO) and further serially diluted to yield various working concentrations for subsequent incubation assays. Icariin with a purity of 99% (batch number: 1810210) was supplied by Yangtze River Pharmaceutical Group; DMSO was also used to prepare its stock solution for serial dilution, and the highest usable concentration in incubation systems was capped at 50 μM owing to solubility saturation at concentrations exceeding this value. As for Compound B, consistent with the requirements of its drug development phase, relevant details including its purity, batch number, manufacturer, storage conditions, physicochemical characteristics, chemical structure and origin are not available for disclosure in this article.

### 1.2 Core Incubation Reagents

0.1 M PBS buffer (pH 7.2–7.4, Jiangsu Kaigi Biotechnology Co., Ltd.); $MgCl_2$ stock solution (hexahydrate, Xilong Chemical); NADPH regenerating cofactor stock (MedchemExpress);; Chromatographic-grade formic acid (Merck), acetonitrile & methanol (TEDIA); DMSO solvent vehicle (Sinopharm Chemical Reagent Co., Ltd.); Internal standards: Verapamil (icariin phase I/II), chlorzoxazone (oroxylin A), caffeine, caffine (compound B).

### 1.3 Hepatic Microsomes

Male human, Sprague-Dawley rat, ICR/CD-1 mouse, beagle dog, cynomolgus monkey liver microsomes (Shanghai Ruide Liver Disease Research Co., Ltd.). Microsome protein concentration was quantified prior to incubation and diluted to target final concentrations.

### 1.4 CYP450 Probe Substrates & Reference Inhibitors

CYP isoform-specific probe substrates, authentic metabolite standards and positive control inhibitors were sourced from Dalian Meilun, Cayman Chemical, TRC, and National Institutes for Food and Drug Control (NIFDC). Full probe panel: phenacetin (CYP1A2), bupropion (CYP2B6), amodiaquine (CYP2C8), diclofenac (CYP2C9), S-mephenytoin (CYP2C19), dextromethorphan (CYP2D6), testosterone & midazolam (dual CYP3A4 marker substrates: CYP3A4(C2) testosterone 6β-hydroxylation; CYP3A4(M) midazolam 1-hydroxylation).

### 1.5 Instrumentation

UPLC System: ACQUITY UPLC I-Class Plus coupled to Xevo TQ-XS triple quadrupole mass spectrometer (Waters), MassLynx V4.2 control software, Analytical balance (Mettler-Toledo XS105DU, 0.01 mg readout), Thermostatted water bath (37 °C constant temperature), Vortex mixer (G560E), High-speed refrigerated centrifuge (Hitachi CT15RE, 15000 rpm), -20 °C and -80 °C medical low-temperature refrigerators (Panasonic), GraphPad Prism 8.0 software for kinetic curve fitting and $IC_{50}$ nonlinear regression

## 2. Phase I NADPH-Dependent Microsomal Metabolic Stability Assay

### 2.1 *In Vitro* Incubation Conditions

A series of concentration gradients was applied for the test compounds in the incubation system. For Oroxylin A, eight gradient concentrations including 0.1, 0.2, 0.5, 1, 2, 5, 10, and 30 μM were adopted universally for all eight CYP isoforms tested. For Icariin, the employed concentration gradient ranged from 0.1 μM to 50 μM, covering 0.1, 0.2, 0.5, 1, 2, 5, 10, 20, and 50 μM, with 50 μM defined as the upper solubility cutoff due to the observed precipitation at concentrations exceeding 50 μM. As the comparator compound, Compound B was tested with different concentration gradients for distinct CYP isoforms: a general gradient of 0.1, 0.5, 1, 5, 10, 50, and 100 μM was used for common CYP isoforms, while a dedicated gradient of 0.1, 0.5, 1, 2, 5, 10, and 15 μM was applied specifically for CYP2C19, and another specialized gradient consisting of 0.01, 0.1, 0.2, 0.5, 1, 5, and 10 μM was adopted for CYP3A4(C2). The final incubation system was constituted of 0.05 M PBS (pH 7.2–7.4), 5.0 mM MgCl, 1.0 mM NADPH cofactor, and 0.2 mg microsomal protein/mL liver microsomes derived from humans, rats, mice, dogs, and monkeys, with the final working concentrations of test compounds set as 10 μM for Oroxylin A, 5 μM for Icariin, and 0.5 μM for Compound B. All incubation mixtures were pre-warmed at 37 °C for 5 min, and the metabolic reactions were subsequently initiated by the addition of NADPH. The reactions were incubated for scheduled durations of 0, 2, 5, 15, 30, 45, 60, and 120 min, with the final organic solvent content of the entire incubation system strictly controlled at ≤1% to eliminate solvent interference. A blank control group without test compounds was incubated for 120 min under identical conditions for comparison. Upon completion of incubation, the reactions were immediately terminated by adding ice-cold internal standard working solution. The resulting samples were vortexed for 5 min, centrifuged at 15000 rpm, 4 °C for 10 min, and supernatant fractions were transferred to UPLC vials for MRM quantification.

### 2.2 Metabolic Kinetic Calculations

The residual substrate fraction, defined as the relative remaining percentage of the substrate, was calculated as the ratio of the analyte peak area ratio (analyte/IS) at time t to that of the initial sample at t = 0. The first-order elimination constant (k) was determined via linear regression based on the equation ln(Ct) = −kt + ln(C0), where Ct represents the substrate concentration at time t and C0 refers to the initial substrate concentration at t = 0; specifically, k (units: ×10 min) corresponds to the negative slope of the curve plotted with the natural logarithm of residual substrate concentration against incubation time. The metabolic half-life (t) was calculated using the formula t = ln2 / k. *In vitro* intrinsic clearance ($CL_{int}$, mL·min·mg protein) was determined according to the equation

$$CL_{int} = \frac{k \times V}{C_{protein}}$$

where V = total incubation volume (0.2 mL); $C_{microsomal\ protein}$ = total microsomal protein mass in the incubation (0.04 mg, from 10 μL of 4 mg/mL microsome stock).

## 3. Human CYP450 Isoform Inhibition Assay

### 3.1 *In Vitro* Incubation Conditions

The standard 200 μL *in vitro* incubation matrix consisted of 0.05 M PBS (pH 7.2–7.4), 5 mM MgCl, 1 mM NADPH cofactor, and human liver microsomes at a final protein concentration of 0.2 mg/mL, supplemented with fixed isoform-specific probe substrates and serially diluted test compounds, while the organic solvent volume fraction in the system was strictly controlled at ≤1% to eliminate potential interference with enzymatic activity; all reaction mixtures were pre-incubated at 37 °C for 5 min, and metabolic reactions were initiated by subsequent addition of NADPH. Differentiated serial concentration gradients were applied to different test compounds for CYP isoform inhibition assessment: Oroxylin A adopted a universal gradient of 0.1, 0.2, 0.5, 1, 2, 5, 10, and 30 μM for all eight tested CYP isoforms; Icariin was tested at concentrations of 0.1, 0.2, 0.5, 1, 2, 5, 10, 20, and 50 μM, with 50 μM set as the upper solubility cutoff due to visible precipitation at higher concentrations; for the comparator Compound B, a general gradient of 0.1, 0.5, 1, 5, 10, 50, and 100 μM was used for common CYP isoforms, a dedicated gradient of 0.1, 0.5, 1, 2, 5, 10, and 15 μM was specifically configured for CYP2C19, and a refined gradient of 0.01, 0.1, 0.2, 0.5, 1, 5, and 10 μM was applied for CYP3A4(C2). All reaction mixtures were pre-incubated at 37 °C for 5 min, and the metabolic reactions were initiated by the addition of 1 mmol/L NADPH. A blank control group without test compounds was established under identical incubation conditions for comparison. To match the metabolic characteristics of different CYP isoforms, isoform-specific incubation durations were applied: 20 min for phenacetin (CYP1A2 substrate), 30 min for bupropion hydrochloride (CYP2B6 substrate) and S-mephenytoin (CYP2C19 substrate), 10 min for amodiaquine (CYP2C8 substrate) and testosterone (CYP3A4(C2) substrate), 15 min for diclofenac (CYP2C9 substrate) and dextromethorphan (CYP2D6 substrate), and 5 min for midazolam (CYP3A4(M) substrate). Upon completion of incubation, the reactions were immediately terminated by adding ice-cold internal standard working solution. The samples were subsequently vortexed for 5 min, centrifuged at 15000 rpm, 4 °C for 10 min, and supernatant fractions were transferred to UPLC vials for MRM quantification.

### 3.2 $IC_{50}$ Calculations

The inhibitory percentage corresponding to each test compound concentration was calculated using the formula:

$$\text{Inhibition (\%)} = \left(1 - \frac{\text{Metabolite peak ratio (+ test compound)}}{\text{Metabolite peak ratio (blank control)}}\right) \times 100$$

The dose–response curves were fitted by nonlinear regression using GraphPad Prism 8.0 to determine the $IC_{50}$ values for each CYP isoform, and compounds with an $IC_{50}$ less than 10 μM were considered to exhibit no significant inhibitory activity against the corresponding CYP isoform.

## 4. UPLC-MS/MS Analytical Methods (SI Standardized Full Chromatography & MS MRM Parameters)

All analytes (parent test compounds, CYP probe metabolites, internal standards) were quantified via dedicated UPLC-MS/MS MRM ESI positive/negative ion methods; cone voltage, collision energy, and diagnostic ion transitions were documented in supplementary Table S10. Representative column stationary phases: Agilent Extend-C18 (1.8 μm), Waters ACQUITY BEH C18 (1.7 μm), with matching guard pre-columns. Mobile phase modifiers: 0.1% formic acid for positive ion mode; 0.1% formic acid + 5 mM ammonium acetate for negative ion detection. All analytical methods were fully validated for linear range (20–3000 nM), LLOQ, intra/inter-batch precision (%CV <15%), and accuracy (%bias±15%) prior to sample analysis.

## 5. Data Visualization & Statistical Analysis

All metabolic stability curves (relative residual substrate vs incubation time), first-order ln-linear kinetic plots, and CYP inhibition dose-response curves were generated in GraphPad Prism 8.0. Kinetic parameters (k, t1/2, $CL_{int}$) and $IC_{50}$ values were reported as mean values of n=3 biological replicates with standard deviation where applicable. Interspecies metabolic clearance comparisons were interpreted based on relative CLint magnitude; $IC_{50}$ thresholds for inhibitory classification: potent ($IC_{50}$ < 1 μM), moderate (1–20 μM), weak (20–50 μM), negligible (>50 μM).

## Supplementary Tables

### Table S1. Details of downstream datasets.

| Properties | Endpoints | Molecules | | Units |
|---|---|---|---|---|
| Absorption | Caco2 | 1275 | Regression | 10^-6 cm/s |
| | PAMPA | 5447 | Classification | / |
| | Pgp | 1911 | Classification | / |
| Distribution | Fu | 906 | Regression | / |
| | VDss | 1164 | Regression | L/kg |
| Metabolism | CYP1A2 inhibitor | 3159 | Classification | / |
| | CYP1A2 substrate | 2341 | Classification | / |
| | CYP2C19 inhibitor | 3200 | Classification | / |
| | CYP2C19 substrate | 2298 | Classification | / |
| | CYP2C9 inhibitor | 3106 | Classification | / |
| | CYP2C9 substrate | 2604 | Classification | / |
| | CYP2D6 inhibitor | 3205 | Classification | / |
| | CYP2D6 substrate | 2604 | Classification | / |
| | CYP3A4 inhibitor | 2967 | Classification | / |
| | CYP3A4 substrate | 2965 | Classification | / |
| | HLM | 3918 | Regression | / |
| Excretion | Clearance | 1294 | Regression | L/h/kg |
| | MRT | 1226 | Regression | h |
| | HL | 1281 | Regression | h |
| Toxicity | AMES | 9140 | Classification | / |
| | DILIst | 1279 | Classification | / |

Unless otherwise stated, all labels are derived from human assay results.

**Table S2. Test Performance of different models on downstream classification tasks.**

| Dataset | CYP1A2_sub | CYP2C9_sub | CYP2C19_sub | CYP2D6_sub | CYP3A4_sub | PAMPA |
|---|---|---|---|---|---|---|
| **RF** | **87.48±2.95** | **81.22±3.12** | 86.32±3.80 | **87.32±3.30** | 85.35±0.19 | 80.36±0.51 |
| **SVM** | 86.46±3.20 | 72.61±3.13 | 80.92±3.29 | 86.37±2.91 | 85.84±0.84 | 80.73±0.78 |
| **XGBoost** | **87.67±1.39** | 79.44±3.77 | 85.37±3.29 | 86.75±3.58 | 85.76±0.35 | **81.77±0.82** |
| **Logistics** | 84.56±2.70 | 73.64±2.42 | 79.61±3.29 | 83.71±1.86 | 84.49±0.43 | 76.96±1.05 |
| **Bayes** | 61.42±4.41 | 56.08±5.43 | 62.02±3.27 | 62.72±3.18 | 73.84±0.26 | 71.58±0.3 |
| **ChemProp** | 84.43±0.41 | 71.02±0.14 | 82.43±0.73 | 84.78±0.52 | **86.19±0.23** | 81.77±0.25 |
| **MEGA-CL** | 85.38±1.28 | **81.53±4.34** | 79.77±6.52 | **87.79±1.6** | **88.63±1.58** | **82.55±1.73** |

| Dataset | CYP1A2_in | CYP2C9_in | CYP2C19_in | CYP2D6_in | CYP3A4_in | P-gp | DIList |
|---|---|---|---|---|---|---|---|
| **RF** | **94.01±0.63** | 89.76±2.78 | 88.22±0.22 | 89.66±1.08 | 91.32±1.79 | **87.85±4.09** | 65.11±1.64 |
| **SVM** | 92.99±1.38 | 89.01±2.48 | **89.72±4.16** | 89.27±1.62 | **92.04±0.95** | **88.02±2.68** | **66.63±2.35** |
| **XGBoost** | 93.72±1.24 | 88.39±1.98 | 88.14±0.69 | 90.11±1.35 | 91.72±1.43 | 86.59±4.11 | **65.43±0.96** |
| **Logistics** | 93.20±0.76 | **89.97±2.66** | 87.23±1.68 | 90.45±0.86 | 91.53±0.55 | 85.39±1.27 | 58.61±2.79 |
| **Bayes** | 82.45±2.26 | 65.62±3.15 | 61.48±1.57 | 59.03±7.11 | 57.52±4.60 | 65.94±3.96 | 65.32±1.95 |
| **ChemProp** | 90.98±0.37 | **92.09±0.06** | 89.32±0.25 | **92.13±0.21** | 91.05±0.36 | 87.7±0.61 | 61.96±2.00 |
| **MEGA-CL** | **94.38±1.83** | 88.2±1.06 | **90.48±0.88** | **92.25±1.83** | **92.89±1.28** | 84.95±0.52 | 64.10±2.79 |

Mean ± SD over three repeat runs.The two best results are in bold.

**Table S3. Test Performance of different models on downstream regression tasks.**

| Dataset | Caco2 | VDss | HL | Fu | MRT | CL | HLM |
|---|---|---|---|---|---|---|---|
| **RF** | **0.42±0.020** | 1.71±0.110 | 10.87±0.910 | 0.24±0.016 | **7.86±0.357** | **9.01±0.970** | 0.68±0.03 |
| **SVM** | 0.43±0.010 | 1.92±0.040 | 10.92±0.740 | 0.25±0.014 | **8.00±0.419** | 9.12±0.854 | 0.67±0.03 |
| **XGBoost** | **0.39±0.010** | 1.75±0.160 | 11.69±1.500 | **0.24±0.010** | 8.09±0.722 | 9.25±0.843 | **0.66±0.03** |
| **Logistics** | 0.52±0.040 | 2.19±0.230 | 17.53±0.610 | 0.31±0.014 | 12.37±0.735 | 13.30±0.612 | 1.27±0.04 |
| **ChemProp** | 0.47±0.002 | **1.57±0.020** | **10.43±0.050** | 0.25±0.005 | 8.21±0.246 | 10.59±0.039 | 0.73±0.00 |
| **MEGA-CL** | 0.45±0.060 | **0.92±0.170** | **10.38±1.670** | **0.21±0.021** | 8.13±1.445 | **7.59±0.953** | **0.53±0.00** |

Mean ± SD over three repeat runs.The two best results are in bold.

**Table S4. Phase I *In Vitro* Metabolic Kinetics Parameters of Oroxylin in Liver Microsomes from Different Species.**

| Microsomes | $C_{(protein)}$(mg/mL) | $\lambda(10^{-3}min^{-1})$ | $t_{1/2}$(min) | $CL_{in\ vitro}(\mu L\cdot min^{-1}mg\ protein^{-1})$ |
|---|---|---|---|---|
| Dog | 0.2 | 2.8 | 247.5 | 14.0 |
| Human | 0.2 | 6.6 | 105.0 | 33.0 |
| Rat | 0.2 | 8.0 | 86.63 | 40.0 |
| Monkey | 0.2 | 13.9 | 49.86 | 69.5 |
| Mouse | 0.2 | 19.3 | 35.91 | 96.5 |

**Table S5. Phase I *In Vitro* Metabolic Kinetics Parameters of compound B in Liver Microsomes from Different Species.**

| Microsomes | $C_{(protein)}$(mg/mL) | $\lambda(10^{-3}min^{-1})$ | $t_{1/2}$(min) | $CL_{in\ vitro}(\mu L\cdot min^{-1}mg\ protein^{-1})$ |
|---|---|---|---|---|
| Dog | 0.2 | 6.0 | 115.5 | 30.0 |
| Human | 0.2 | 6.3 | 110.0 | 31.5 |
| Monkey | 0.2 | 9.2 | 75.33 | 46.0 |
| Rat | 0.2 | 5.2 | 133.3 | 26.0 |
| Mouse | 0.2 | 6.7 | 103.4 | 33.5 |

**Table S6. Phase I *In Vitro* Metabolic Kinetics Parameters of Icariin in Liver Microsomes from Different Species.**

| Microsomes | $C_{(protein)}$(mg/mL) | $\lambda(10^{-3}min^{-1})$ | $t_{1/2}$(min) | $CL_{in\ vitro}(\mu L\cdot min^{-1}mg\ protein^{-1})$ |
|---|---|---|---|---|
| Dog | 0.2 | 0.60 | 1155 | 3.0 |
| Human | 0.2 | 0.20 | 3465 | 1.0 |
| Monkey | 0.2 | 0.05 | 13860 | 0.2 |
| Rat | 0.2 | 1.40 | 495.0 | 7.0 |
| Mouse | 0.2 | 0.30 | 2310 | 1.5 |

**Table S7. Test Performance of different models on benchmarks from Pharma-bench.**

| Dataset | LogD | Solubility | PPB | CYP 3A4 | CYP 2C9 | CYP 2D6 | HLMC | RLMC | MLMC |
|---|---|---|---|---|---|---|---|---|---|
| **RF** | 1.249 | 0.918 | 0.204 | 16.54 | 18.471 | 19.826 | 0.813 | 0.958 | 0.987 |
| **XGBoost** | 1.071 | 0.832 | 0.186 | 16.123 | 17.582 | 20.685 | 0.647 | 0.819 | 0.844 |
| **CMPNN** | 0.807 | 0.858 | 0.236 | 16.701 | 18.377 | 21.612 | 0.921 | 0.939 | 1.13 |
| **FPGNN** | 0.838 | **0.747** | 0.179 | 15.606 | **16.933** | 19.948 | 0.604 | 0.716 | 0.774 |
| **DHTNN** | 0.912 | 0.828 | 0.235 | 16.156 | 17.449 | 20.467 | 0.729 | 0.915 | 0.926 |
| **KANO** | 0.766 | 0.772 | 0.185 | **15.307** | 17.35 | **19.385** | 0.554 | 0.762 | 0.767 |
| **MPG** | 0.758 | 0.758 | **0.17** | **14.376** | 17.417 | 20.012 | **0.541** | 0.685 | **0.723** |
| **Unimol** | 0.745 | **0.707** | 0.179 | 15.895 | 17.774 | 20.695 | 0.613 | **0.651** | 0.824 |
| **Trans-M** | **0.737** | 0.834 | **0.172** | 15.867 | 18.08 | 20.664 | 0.567 | 0.677 | 0.744 |
| **MEGA-CL** | **0.70 ± 0.09** | 0.803±0.01 | 0.18 ± 0.01 | 16.17±0.49 | **16.39±0.57** | **17.60±0.41** | **0.53±0.01** | **0.59±0.10** | **0.63 ± 0.033** |

Results for MEGA-CL are presented as mean±SD.The two best results are in bold.

**Table S8. External Validation: Molecular Overview and Predictive Performance.**

| Name | Predicted HLMC | Observed HLMC | Fold error | Molecular weight | LogP |
|---|---|---|---|---|---|
| Capmatinib | 72.54 | 61.85 | 1.17 | 412.43 | 3.43 |
| Pemigatinib | 25.65 | 25.40 | 1.01 | 487.51 | 3.66 |
| Selpercatinib | 25.69 | 34.00 | 1.32 | 525.61 | 3.28 |
| Tandutinib | 11.61 | 22.03 | 1.90 | 562.72 | 5.03 |
| Veliparib | 20.15 | 22.23 | 1.10 | 244.30 | 1.26 |
| Ensartinib | 18.56 | 42.00 | 2.26 | 561.45 | 4.72 |
| Niraparib | 8.57 | 24.10 | 2.81 | 320.40 | 2.59 |
| Lenvatinib | 47.93 | 24.20 | 1.98 | 426.86 | 4.07 |
| Sorafenib | 52.48 | 30.80 | 1.70 | 464.83 | 5.55 |
| Nebivolol | 21.85 | 39.60 | 1.81 | 405.44 | 2.36 |
| Cediranib | 40.28 | 31.80 | 1.27 | 450.51 | 5.22 |
| Veliparib | 23.32 | 22.20 | 1.05 | 244.30 | 1.26 |
| Olaparib | 11.61 | 18.60 | 1.60 | 434.47 | 2.35 |
| Infigratinib | 20.15 | 24.00 | 1.19 | 560.49 | 5.35 |
| Zorifertinib | 18.56 | 33.50 | 1.80 | 459.91 | 4.31 |
| Talazoparib | 8.57 | 9.59 | 1.12 | 380.36 | 2.63 |

| Crenolanib | 27.49 | 26.79 | 1.03 | 443.55 | 3.92 |
|---|---|---|---|---|---|
| Ertugliflozin | 18.71 | 12.00 | 1.56 | 566.00 | 0.71 |

Units for HLM $Cl_{int}$ were converted to $\mu L \cdot min^{-1} \cdot mg^{-1}$ protein). Molecular weights and logP were computationally derived.

**Table S9. Incubation Matrix Formulation of Phase I NADPH-Dependent Microsomal Metabolic Stability Assay.**

| Component | Final Concentration in System |
|---|---|
| 0.1 M PBS (pH 7.2–7.4) | 0.05 M |
| 20 mM $MgCl_2$ | 5.0 mM |
| Test compound working solution | Oroxylin A 10 μM; Icariin 5 μM; Compound B 0.5 μM |
| 10 mM NADPH cofactor | 1.0 mM |
| Liver microsomes (human/rat/mouse/dog/monkey) | 0.2 mg microsomal protein/mL |

**Table S10. Fixed Final Concentrations of CYP Probe Substrates.**

| CYP Isoform | Marker Reaction | Probe Final Concentration | Incubation Duration |
|---|---|---|---|
| CYP1A2 | Phenacetin O-deethylation | 60 μM | 20 min |
| CYP2B6 | Bupropion hydroxylation | 140 μM | 30 min |
| CYP2C8 | Amodiaquine N-desethylation | 10 μM | 10 min |
| CYP2C9 | Diclofenac 4-hydroxylation | 6 μM | 15 min |
| CYP2C19 | S-mephenytoin 4-hydroxylation | 20 μM | 30 min |
| CYP2D6 | Dextromethorphan O-demethylation | 5 μM | 15 min |
| CYP3A4(C2) | Testosterone 6β-hydroxylation | 50 μM | 10 min |
| CYP3A4(M) | Midazolam 1-hydroxylation | 10 μM | 5 min |

**Table S11. UPLC-MS/MS MRM detection parameters for three test compounds and six CYP probe substrates**

Instrument: ACQUITY UPLC I-Class Plus coupled with Xevo TQ-XS triple quadrupole mass spectrometer (Waters); ion source: electrospray ionization (ESI); detection mode: multiple reaction monitoring (MRM); column temperature: 40 °C; injection volume: 1–2 μL.

| Compound Name | Ion Mode | Precursor Ion (m/z) | Quantifier Ion (m/z) | Cone Voltage (V) | Collision Energy (eV) | Retention Time (min) |
|---|---|---|---|---|---|---|
| Oroxylin A | ESI+ | 285.1 | 270.1 | 35 | 22 | 5.2 |
| Icariin | ESI+ | 677.2 | 531.2 | 45 | 30 | 4.8 |
| Compound B | ESI+ | 301.1 | 286.1 | 32 | 20 | 5.5 |
| Phenacetin | ESI+ | 180.1 | 138.1 | 25 | 18 | 3.1 |
| Amodiaquine | ESI+ | 356.1 | 285.1 | 40 | 25 | 3.8 |
| Diclofenac | ESI− | 310.0 | 266.0 | 29 | 19 | 4.5 |
| S-mephenytoin | ESI− | 233.0 | 190.0 | 35 | 15 | 3.3 |
| Dextromethorphan | ESI+ | 258.1 | 132.9 | 15 | 30 | 3.6 |
| Testosterone | ESI+ | 305.1 | 269.06 | 36 | 12 | 4.2 |
| Midazolam | ESI+ | 342.1 | 202.8 | 10 | 24 | 3.0 |

## Supplementary Figures

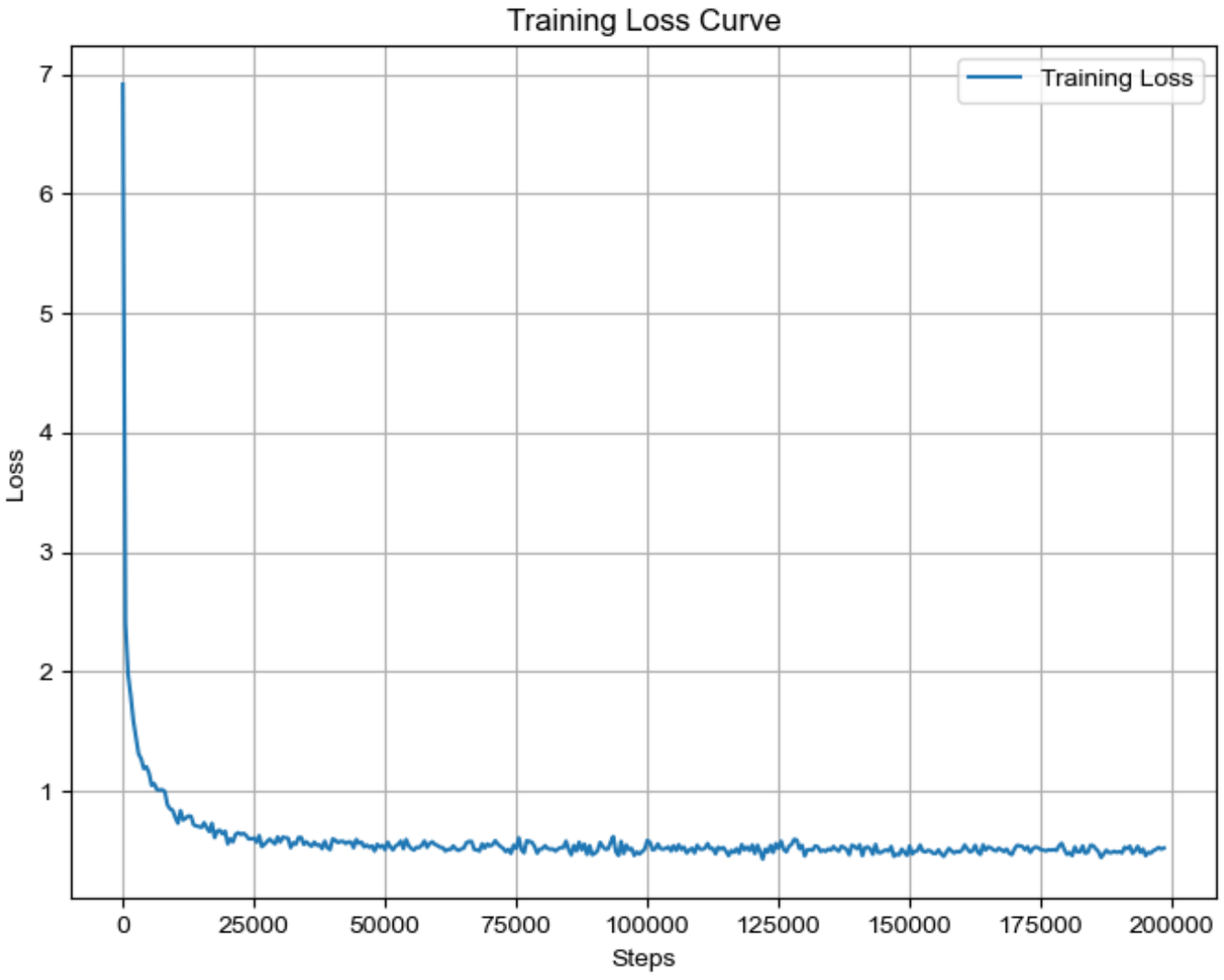

**Figure S1. Training loss curve recorded during model pre-training.** Training loss was plotted as a function of training steps throughout the pre-training process, which spanned 100 epochs with a batch size set to 512. For robust self-supervised molecular representation learning, a large-scale pre-training dataset was curated from the PubChem database: roughly $1.7 \times 10^8$ raw molecular entries were initially retrieved, followed by strict filtering procedures to eliminate duplicate records, invalid chemical structures, and disconnected moieties including inorganic salts and auxiliary solvents. After purification, a final dataset containing 100 million unique unlabeled molecular SMILES sequences was obtained and employed for pre-training.

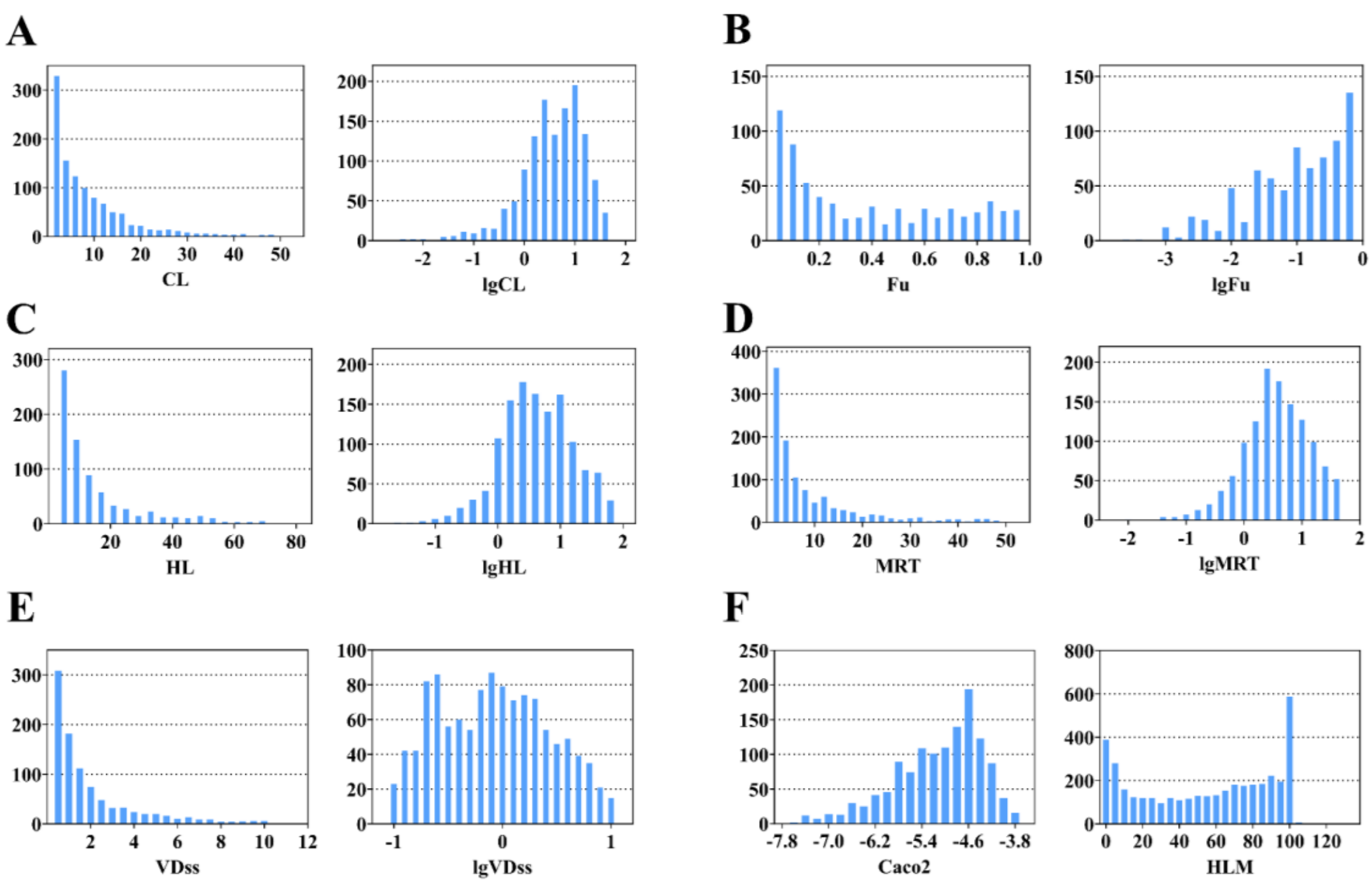

**Figure S2. Distribution of the regression task dataset. A** Clearance (CL) and its logarithmic transformation (lgCL). **B** Fraction unbound (Fu) and its logarithmic transformation (lgFu). **C** Half-life (HL) and its logarithmic transformation (lgHL). **D** Mean residence time (MRT) and its logarithmic transformation (lgMRT). **E** Volume of distribution at steady state (VDss) and its logarithmic transformation (lgVDss).Blue bars represent the frequency distribution of each parameter. **F** Caco2 and HLM in original scale. The x-axis shows the parameter values (original scale) or their base-10 logarithmic values (log scale), the y-axis indicates the frequency count.

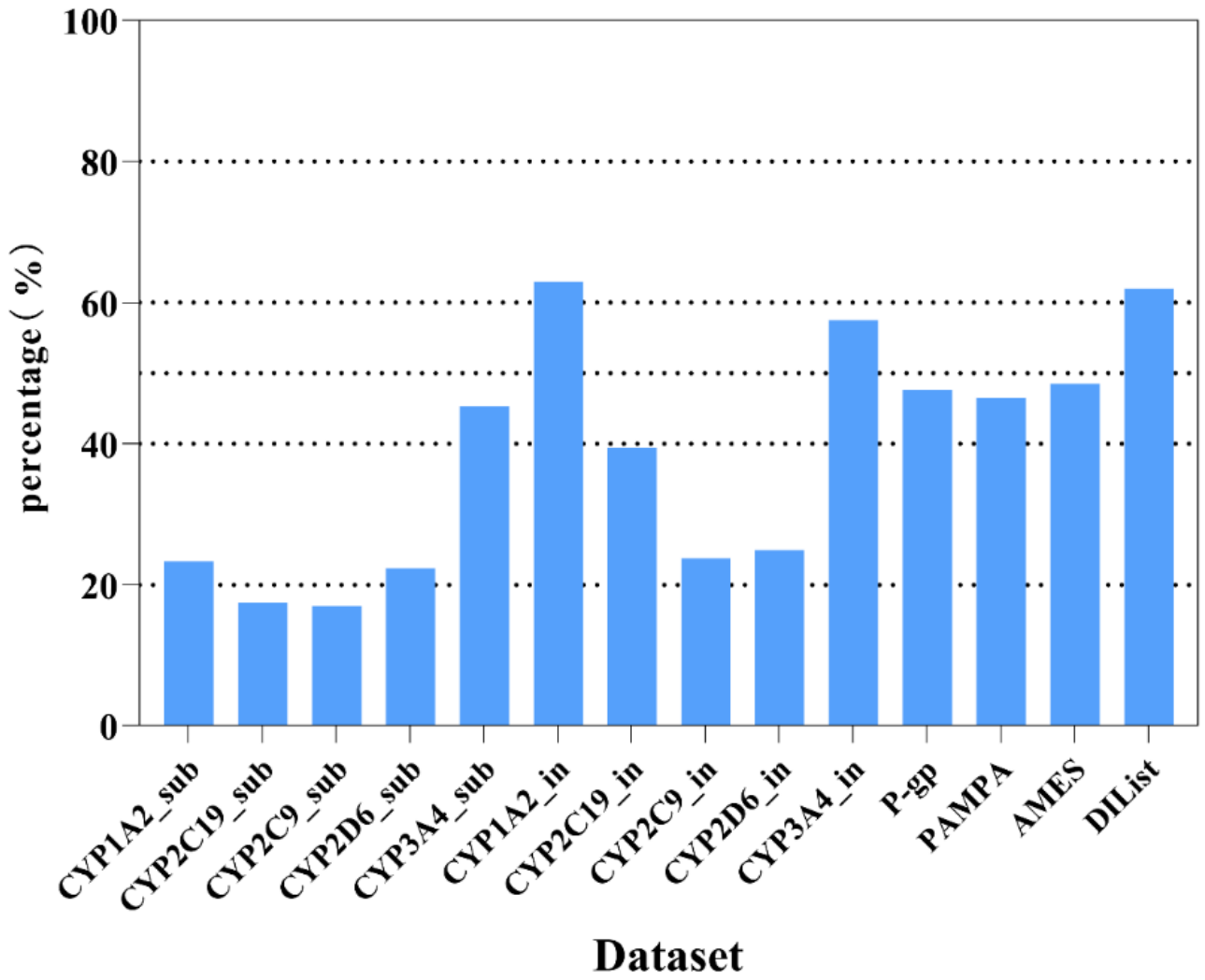

**Figure S3. Distribution of the classification task dataset.** The percentage of molecules labeled as positive in each classification tasks.